%% file: main.tex
\documentclass[10pt,twocolumn,letterpaper]{article}

\usepackage[pagenumbers]{cvpr} 

\input{preamble}

\definecolor{cvprblue}{rgb}{0.21,0.49,0.74}
\usepackage[pagebackref,breaklinks,colorlinks,allcolors=cvprblue]{hyperref}

\def\paperID{807} 
\def\confName{CVPR}
\def\confYear{2025}

\def\eg{\emph{e.g.}}
\def\ie{\emph{i.e.}}

\def\etal{\emph{et al.\ }}

\title{Masked Clustering Prediction for Unsupervised Point Cloud Pre-training}

\author{  
\thanks{Indicates equal contribution.}~Bin Ren\textsuperscript{1,2}, 
\footnotemark[1]~Xiaoshui Huang\textsuperscript{3}, 
Mengyuan Liu\textsuperscript{4},
Hong Liu\textsuperscript{4},
Fabio Poiesi\textsuperscript{5},
Nicu Sebe\textsuperscript{2}, 
\thanks{Corresponding to Guofeng Mei (gfmeiwhu@gmail.com) }~Guofeng Mei$^5$ \\
\normalsize{
\textsuperscript{1}University of Pisa, 
\textsuperscript{2}University of Trento, 
\textsuperscript{3}Shanghai Jiao Tong University,
\textsuperscript{4}Peking University, 
\textsuperscript{5}Fondazione Bruno Kessler}\\
}

\begin{document}
\maketitle
\input{secs/0_abstract}    
\input{secs/1_intro}
\input{secs/2_related}
\input{secs/3_method}
\input{secs/4_experiments}
\input{secs/5_conclusion}

\input{secs/X_suppl}

{
    \small
    \bibliographystyle{ieeenat_fullname}
    \bibliography{main}
}

\end{document}

%% file: preamble.tex
\usepackage[dvipsnames]{xcolor}
\usepackage{multirow}
\usepackage{amsmath}
\usepackage{algorithm}%
\usepackage{algorithmicx}%
\usepackage{algpseudocode}
\usepackage{amsfonts}
\usepackage{graphicx}
\usepackage{bm,color}
\usepackage{colortbl}
\usepackage{rotating}
\usepackage{array}
\usepackage[percent]{overpic}
\usepackage{listings}

\newcommand{\ourmethod}{MaskClu\xspace}

\definecolor{light-gray}{gray}{0.5}
\definecolor{pretty-blue}{RGB}{0, 113, 188}
\definecolor{rowcolor}{gray}{.95} 
\definecolor{linecolor}{gray}{.895} 

\makeatletter
\newcommand{\compressalg}{
  \setlength{\@tempdima}{\algorithmicindent}
  \renewcommand{\algorithmicindent}{\@tempdima}%
  \setlength{\itemsep}{0pt}
  \setlength{\parskip}{0pt}
  \setlength{\parsep}{0pt}
}
\makeatother

\usepackage{etoolbox}
\makeatletter
\AfterEndEnvironment{algorithm}{\let\@algcomment\relax}
\AtEndEnvironment{algorithm}{\kern2pt\hrule\relax\vskip3pt\@algcomment}
\let\@algcomment\relax
\newcommand\algcomment[1]{\def\@algcomment{\footnotesize#1}}
\renewcommand\fs@ruled{\def\@fs@cfont{\bfseries}\let\@fs@capt\floatc@ruled
  \def\@fs@pre{\hrule height.8pt depth0pt \kern2pt}%
  \def\@fs@post{}%
  \def\@fs@mid{\kern2pt\hrule\kern2pt}%
  \let\@fs@iftopcapt\iftrue}
\makeatother

\definecolor{codeblue}{rgb}{0.25,0.5,0.5}
\definecolor{codekw}{rgb}{0.85, 0.18, 0.50}
\newcommand{\myparagraph}[1]{\vspace{2pt}\noindent{\bf #1}}

%% file: secs/0_abstract.tex
\begin{abstract}
Vision transformers (ViTs) have recently been widely applied to 3D point cloud understanding, with masked autoencoding as the predominant pre-training paradigm. However, the challenge of learning dense and informative semantic features from point clouds via standard ViTs remains underexplored.
We propose \ourmethod, a novel unsupervised pre-training method for ViTs on 3D point clouds that integrates masked point modeling with clustering-based learning. \ourmethod{} is designed to reconstruct both cluster assignments and cluster centers from masked point clouds, thus encouraging the model to capture dense semantic information. 
Additionally, we introduce a global contrastive learning mechanism that enhances instance-level feature learning by contrasting different masked views of the same point cloud. 
By jointly optimizing these complementary objectives, \ie, dense semantic reconstruction, and instance-level contrastive learning. \ourmethod{} enables ViTs to learn richer and more semantically meaningful representations from 3D point clouds.
We validate the effectiveness of our method via multiple 3D tasks, including part segmentation, semantic segmentation, object detection, and classification, where \ourmethod{} sets new competitive results. The code and models will be released at:~\url{https://github.com/Amazingren/maskclu}.
\end{abstract}

%% file: secs/1_intro.tex
\section{Introduction}
\label{sec:intro}
Unsupervised pre-training has proven to be highly effective in enhancing point cloud understanding, sparking the development of numerous unsupervised methods~\cite{mei2022data,eckart2021self,ren2024bringing,liang2024pointmamba,zheng2024point,ma2024shapesplat} specifically tailored for 3D point clouds. These methods have demonstrated remarkable versatility and have been applied across a wide range of domains, including autonomous driving, robotics, virtual/augmented reality, and large-scale 3D scene understanding~\cite{wang2025rolo,miao2024scenegraphloc,ren2021cascaded,jiao2025free,zheng2024deep,li2025pvafn}.

In general, unsupervised point cloud pre-training approaches can be categorized into three main types: contrastive learning~\cite{grill2020bootstrap,li2025scenesplat}, clustering-based learning, and generative learning.
The first category, contrastive learning~\cite{rao2020global,sanghi2020info3d,ma2025scenesplat++}, has shown remarkable success in recent point cloud representation tasks. Its core idea is to align representations with positive examples while pushing them away from negative ones, relying heavily on strong data augmentations and scaling effectively with large datasets.

\begin{figure}[!t]
    \centering
    \includegraphics[width=1.0\linewidth]{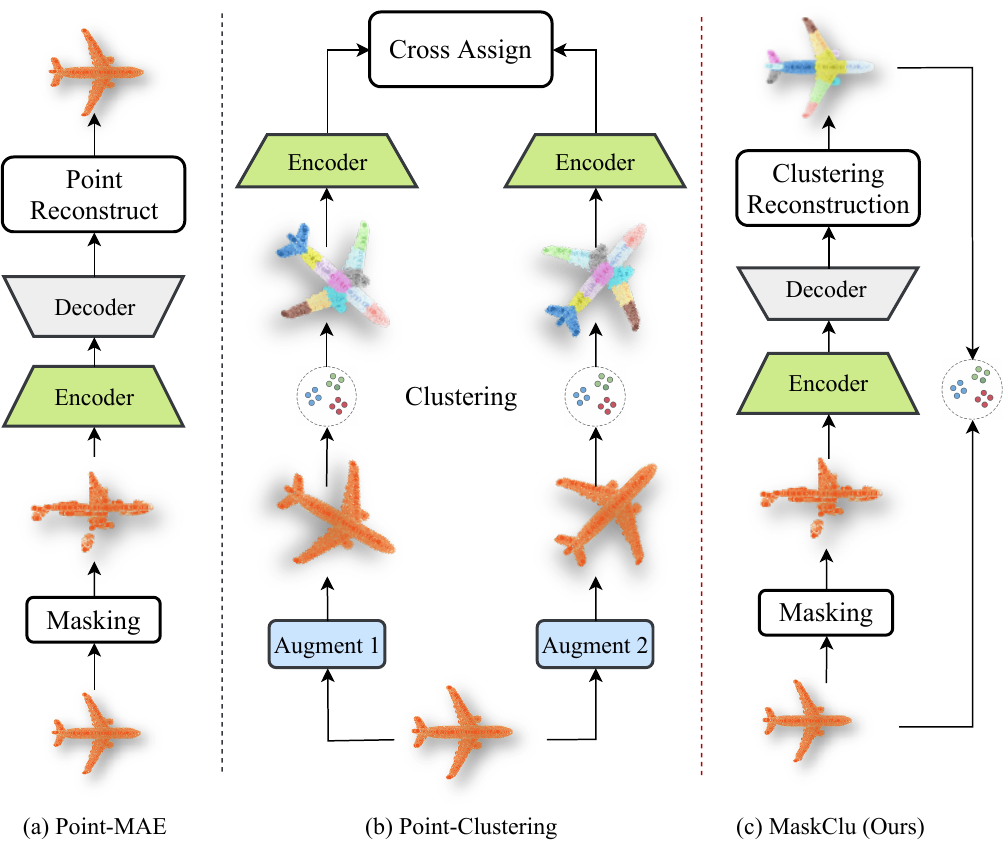}
    \vspace{0mm}
    \caption{The proposed \ourmethod{~}(c) combines the strengths of masked modeling (a) and clustering-based learning (b) to produce more informative representations.}
    \label{fig:teaser}
    \vspace{0mm}
\end{figure}
Clustering-based methods~\cite{mei2022data,mei2024unsupervised}, which require minimal domain knowledge, have gained prominence in the pre-training of both 2D and 3D networks. These approaches group similar feature embeddings and refine network parameters using pseudo-labels generated through clustering. Unlike contrastive learning, which primarily captures intra-object invariance, clustering methods enable the exploration of inter-object similarities~\cite{zhan2020online}.
On the generative side, methods such as autoencoders~\cite{yang2018foldingnet}, adversarial generative networks (GAN)~\cite{sarmad2019rl}, and autoregressive models~\cite{sun2020pointgrow} have been effective in capturing the low-level properties of point clouds. 
In particular, masked autoencoders (MAEs) such as Point-MAE~\cite{liu2022masked}, Point-M2AE~\cite{MA2E} and Point-CMAE~\cite{ren2024bringing} are widely used to pre-train point cloud ViTs by reconstructing masked point patches. Experimental results indicate that masked point cloud modeling surpasses contrastive learning and has emerged as the state-of-the-art approach for unsupervised pretraining of point cloud ViTs.

However, MAEs optimize a point-level reconstruction objective that emphasizes spatial relationships in the learned features rather than capturing semantic information. In contrast, clustering methods focus on higher-level feature similarity, producing more semantically meaningful dense representations. 
This raises an interesting question: \textit{Can we combine the strengths of both masked modeling and clustering learning to achieve better representations?} Moreover, considering that both local and global shape information are essential in 3D point cloud analysis, \ie, the global information captures the overall structure, while the local information provides the fine-grained details crucial for a detailed understanding. 

These insights led us to develop \ourmethod{}, a masked clustering prediction approach for unsupervised point cloud pre-training that combines the benefits of both masked modeling and clustering. Specifically, \ourmethod{} applies a graph convolutional network (GCN) to reconstruct both clustering assignments and cluster centers from masked point clouds, enhancing semantic consistency. 
To further enrich the representations, we incorporate contrastive learning to capture global features effectively.
Fig.~\ref{fig:teaser} highlights the differences between \ourmethod{} and prior SOTA solutions. Unlike PointClustering~\cite{long2023pointclustering}, which clusters based solely on coordinates and supervises via cross-view feature similarity, \ourmethod{} exploits both geometric and feature cues, and learns by reconstructing cluster assignments and centers, yielding more robust and semantic representations.

To validate the effectiveness of \ourmethod, we validate it on 4 benchmarks: 
part segmentation (ShapeNetPart~\cite{chang2015shapenet}), 
semantic segmentation (S3DIS~\cite{armeni_cvpr16}), 
object detection (ScanNet~\cite{dai2017scannet}), 
and classification (ScanObjectNN~\cite{scanobjectnn}). 
\ourmethod{} shows high transferability across dense prediction tasks, achieving competitive results.

Our main contributions are summarized as follows:
\begin{itemize}
    \item We present a novel ViT-based unsupervised pretraining framework for 3D point clouds that unifies masked point modeling with clustering-based representation learning.

    \item We introduce a cluster reconstruction objective that predicts both cluster assignments and centers from masked point clouds, enabling dense semantic feature learning.

    \item We design an instance-level contrastive strategy by contrasting dual masked views to enhance the discriminability of learned representations.

    \item Extensive experiments on multiple benchmarks demonstrate the superiority of our approach over prior state-of-the-art unsupervised methods in dense prediction tasks.
\end{itemize}

%% file: secs/2_related.tex
\section{Related Work}
\label{sec:related-work}
We provide an overview of the key unsupervised learning techniques for point cloud pre-training, categorized into contrastive methods, clustering-based approaches, and generative models.

\noindent\textbf{Contrastive learning} has become a prominent framework for unsupervised point cloud representation learning, aiming to maximize agreement between different augmentations of the same point cloud while minimizing similarity across distinct samples. Initially proposed for 2D images~\cite{chen2020simple,grill2020bootstrap,shu2025earthmind}, it has been effectively extended to 3D data~\cite{huang2021spatio,sanghi2020info3d,rao2020global,xie2020pointcontrast,wang2023zero}. 
For instance, Info3D~\cite{sanghi2020info3d} optimizes mutual information between parts of a point cloud and its transformations to enhance representation quality. PointContrast~\cite{xie2020pointcontrast} was an early milestone, followed by PointDisc~\cite{liu2022pointdisc}, which improved feature consistency via geometric priors. 
MaskPoint~\cite{liu2022masked} introduces a binary classification task to distinguish masked object points from noise, while ReCon~\cite{qi2023contrast} embeds contrastive objectives into generative pipelines to mitigate overfitting in ViT-based methods. 
Multimodal extensions further enrich this line of work: Jing~\etal~\cite{jing2021self} leverage 2D-3D correspondences to boost discriminability, and CLIP-inspired approaches such as ULIP~\cite{xue2023ulip} and PointCLIP~\cite{zhang2022pointclip} demonstrate the benefit of incorporating multimodal supervision~\cite{radford2021clip}. 
Despite these advances, contrastive methods often underperform when applied to 3D point clouds with ViTs.

\noindent\textbf{Clustering learning} is an emerging paradigm in unsupervised learning that discovers latent structures by grouping similar feature representations. The resulting clusters often serve as pseudo-labels to guide representation learning. A seminal work, DeepCluster~\cite{caron2018deepcluster}, applied K-means to 2D image features and iteratively refined representations using cluster-derived pseudo-labels. This idea was extended to 3D by SL3D~\cite{cendra2022sl3d}, which employs even cluster sampling to produce high-quality pseudo-labels for 3D recognition. SoftClu~\cite{mei2022data} and CluRender~\cite{mei2024unsupervised} leverage clustering and differentiable rendering to extract discriminative features. 
More recently, PointClustering~\cite{long2023pointclustering} aligns clusters from different views of the same point cloud to enhance point- and instance-level invariance, offering a robust framework for modeling geometric and semantic properties. While clustering-based methods show strong potential, their use in point cloud ViTs remains in its early stages.

\noindent\textbf{Generative models} in unsupervised learning aim to reconstruct point clouds from latent representations by encoding inputs into a feature space and decoding them back. 
FoldingNet~\cite{yang2018foldingnet} uses a graph-based encoder and a folding decoder to deform a 2D grid into a 3D shape. 
Liu~\etal~\cite{liu2019l2g} extended this with a local-to-global autoencoder that hierarchically captures both local and global structures. 
GAN-based methods like Panos~\etal~\cite{achlioptas2018learning} combined hierarchical Bayesian modeling with GANs to synthesize realistic 3D shapes. 
Recent advances focus on masked reconstruction. 
Inspired by MAE~\cite{he2022masked}, Point-MAE~\cite{pang2022masked} simplifies training via MAE. Point-M2AE~\cite{zhang2022point} adopts a hierarchical design to capture multiscale geometric details, while GD-MAE~\cite{yang2023gdmae} extends visible regions to reconstruct masked areas via a generative decoder. Multimodal integration further enriches representations: PiMAE~\cite{chen2023pimae} applies cross-modal masking to fuse visual and geometric cues. 
TAP~\cite{wang2023take} and Ponder~\cite{huang2023ponder} leverage 2D projections for pretraining. 
Joint-MAE~\cite{guo2023joint} models 2D-3D interactions, and PointGPT~\cite{chen2024pointgpt} extends the GPT framework to point clouds. 
PointDif~\cite{zheng2024point} incorporates diffusion models~\cite{ho2020denoising,liu2023spatio,wei2023diffusion,zhao2024denoising} into MAE, while PointMamba~\cite{liang2024pointmamba} pioneers state space models~\cite{gu2022efficiently} for point cloud pretraining. 
Despite progress, generative models remain sensitive to geometric transformations (\ie, rotation, translation) and struggle to reconstruct point clouds from invariant features~\cite{sanghi2020info3d}, limiting their usefulness for downstream tasks.
Despite these advances, generative models remain sensitive to geometric transformations (\eg, rotation, translation), yielding inconsistent representations~\cite{sanghi2020info3d}. Moreover, reconstructing point clouds from invariant features remains challenging, limiting their applicability to downstream tasks.

\begin{figure*}[!t]
    \centering
    \includegraphics[width=1.0\linewidth]{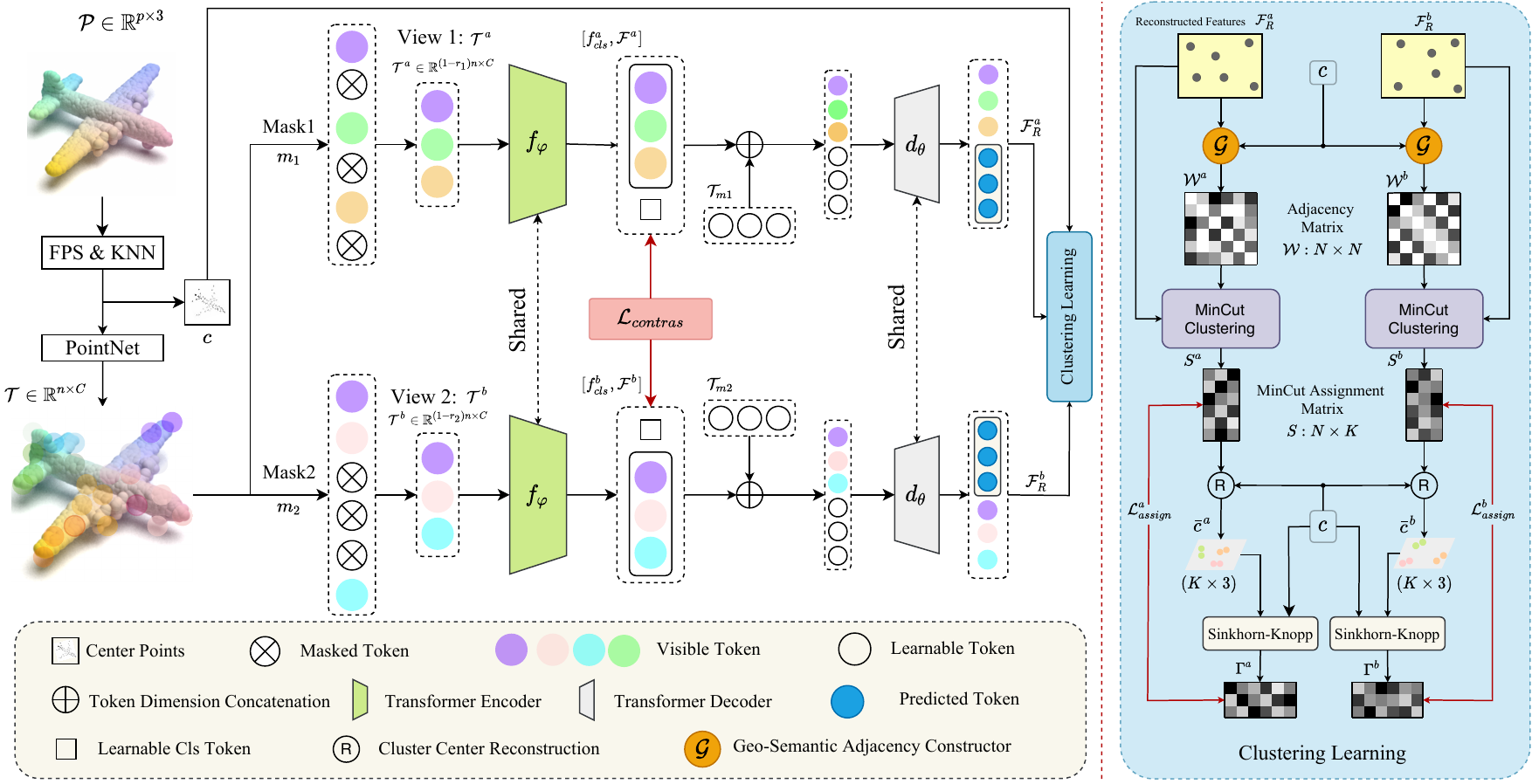}
    \vspace{0mm}
    \caption{
    The architecture of the proposed \ourmethod{} for unsupervised point cloud representation learning integrates local clustering and global contrasting. The former learns dense, fine-grained features, while the latter captures instance-level features.}
    \label{fig:framework}
    \vspace{0mm}
\end{figure*}

%% file: secs/3_method.tex
\section{Methodology}
\label{sec:methodology}
\subsection{Preliminary: Point Patch Embedding}
\label{subsec:preliminary}
We follow Point-BERT~\cite{PointBERT} by first dividing the input point cloud ($\mathcal{P} \in \mathbb{R}^{H \times 3}$, where $H$ denotes the number of points) into irregular point patches using Farthest Point Sampling (i.e., $\operatorname{FPS(\cdot)}$) and K-Nearest Neighbors (i.e., $\operatorname{KNN(\cdot)}$). This process outputs $N$ center points $\mathcal{C}$ ($\mathcal{C} = \operatorname{FPS}(\mathcal{P}), \mathcal{C} \in \mathbb{R}^{N \times 3}$) and the corresponding neighbor points $\mathcal{P}_i$ ($\mathcal{P}_i = \operatorname{KNN}(\mathcal{P}, \mathcal{C}), \mathcal{P} \in \mathbb{R}^{N \times k \times 3}$) for each center point.
Finally, a lightweight PointNet~\cite{li2018point} (denoted as $\operatorname{PN(\cdot)}$), primarily composed of MLPs, is applied to the point patches to obtain the embedded tokens $\mathcal{T}=\{T_1, T_2,\cdots,T_N\}$, where $\mathcal{T} = \operatorname{PN}(\mathcal{P})$ and $\mathcal{T} \in \mathbb{R}^{N \times d}$, with $d$ representing the feature dimension. These tokens are then sent to a standard ViT encoder.

\subsection{Overview}
\label{subsec:overview}
Our goal is to develop an unsupervised training approch for a standard ViT encoder, \( f_\varphi: \mathbb{R}^{rN \times d} \rightarrow \mathbb{R}^{rN \times d} \), where $r$ represents the mask ratio in the MAE (details of the MAE please refer \textit{Supp. Mat.}). The encoder $f_\varphi$ is designed to effectively learn informative patch features $({\mathcal{F}} = \{f_i = f_\varphi(T_i)\}_{i=1}^N)$ as well as a global representation ${\bm{f}}_{cls}$ of $\mathcal{P}$, ultimately benefiting downstream point cloud tasks.
The overall pipeline of our method \ourmethod{} is illustrated in Fig.~\ref{fig:framework}. We adopted a Siamese network architecture~\cite{grill2020bootstrap,chen2020exploring} to perform both patch-level local masked clustering learning and object-level global contrastive learning during unsupervised pre-taining. 
\ourmethod{} begins by generating two randomly masked views, $\mathcal{T}^a$ and $\mathcal{T}^b$, from the patched features $\mathcal{T}$ using two random masks, $m_a$ and $m_b$. The feature extractor $f_\varphi$ is then applied to obtain the features $[f^a_{cls}, \mathcal{F}^b]$ and $[f^b_{cls}, \mathcal{F}^a]$.
Next, $\mathcal{F}^a$ and $\mathcal{F}^b$, along with the learnable position embedding $\mathcal{T}^{m_a}$ and $\mathcal{T}^{m_b}$ corresponding to the masked patches, are passed to a weight-sharing decoder $d_{\theta}$ that reconstructs the full feature representations ($\mathcal{F}^a_R$ and $\mathcal{F}^b_R$).
These reconstructed features, combined with the center points $\mathcal{C}$, are used for masked clustering learning, enhancing $f_\varphi$’s ability in dense prediction tasks. Meanwhile, the feature pairs $[f^a_{cls}, \mathcal{F}^b]$ and $[f^b_{cls}, \mathcal{F}^a]$ are used for global contrastive learning, strengthening the global representation capabilities of $f_{\varphi}$.

\subsection{Masked Clustering Learning}
\label{subsec:clustering}
Traditional MAEs reconstruct masked points by aligning them with a fixed ground-truth shape~\cite{liu2022masked,ren2024bringing}. However, this ground truth corresponds to a single instance from a broader distribution, making optimization challenging and limiting the learning of rotation-invariant features. Furthermore, object-level MAEs often underperform on dense prediction tasks because their objective prioritizes exact geometry reconstruction over capturing higher-level semantic structures, resulting in poor generalization.
In contrast, clustering can uncover latent structures by grouping similar feature representations, making it a powerful complement to MAE-based approaches. Yet, conventional clustering suffers from ambiguous group assignments, particularly when applied independently to geometric or semantic cues. While prior work has addressed these aspects in isolation, achieving joint geometric and semantic coherence remains an open challenge.

\noindent\textbf{Geo-semantic graph construction}
To improve clustering quality, we incorporate geometric cues alongside semantic features. Using only feature space can miss spatial relationships—for example, symmetrical airplane wings may look similar in feature space but differ in spatial layout. By learning representations that are both semantically meaningful and geometrically consistent, our clustering better reflects the true structure of the scene.
\begin{algorithm}[!t]
\vspace{-1mm}
    \caption{Geo-Semantic Graph Construction}
    \label{alg:geo_semantic_graph} 
    \compressalg 
    \begin{algorithmic}[1]
    \Require Center points $\mathcal{C} \in \mathbb{R}^{N \times 3}$, reconstructed features $\mathcal{F}_{R} \in \mathbb{R}^{N \times D}$, neighbors $k$
    \Ensure Affinity matrix $\mathcal{W} \in \mathbb{R}^{N \times N}$
    
    \State $\text{Dis} \leftarrow \text{cdist}(\mathcal{C}, \mathcal{C})$ \quad \# Pairwise distance matrix
    \State $\text{knn\_indices} \leftarrow \text{topk}(\text{Dis}, k)$ \quad \# Indices of $k$ nearest neighbors
    \State $\mathcal{C} \leftarrow 1 + \mathcal{F}_{R} \cdot \mathcal{F}_{R}^\top$ \quad \# Cosine similarity
    \State $\mathcal{W} \leftarrow \mathcal{C} \odot \text{knn\_mask}$ \quad \# Masked by $k$-nearest neighbors
    \State $\mathcal{W} \leftarrow \exp \left( \overline{\text{Dis}} - \text{Dis} \right) \odot \mathcal{W}$  with $\overline{\text{Dis}}_i = \sum_j\text{Dis}_{ij}$ \quad \quad \# Weighted by pairwise distance
    \State $\mathcal{W} \leftarrow \frac{\mathcal{W} + \mathcal{W}^\top}{2}$ \quad \# Symmetric matrix
    \State $\mathcal{W}[i, i] \leftarrow 0, \quad \forall i$ \quad \# No self-loops
    \State \Return $\mathcal{W}$
    \end{algorithmic}
    \vspace{-1mm}
\end{algorithm}
Given the reconstructed feature matrix $\mathcal{F}_R \in \mathbb{R}^{N \times D}$ and the center point coordinates $\mathcal{C} \in \mathbb{R}^{N \times 3}$ of each point patch, we construct a geo-semantic graph $\mathcal{G} = \{\mathcal{C}, \mathcal{W}\}$ by performing Alg.~\ref{alg:geo_semantic_graph}, where $\mathcal{W}$ is an affinity matrix. Focusing on predetermined anchors, we build $\mathcal{W}$ by calculating Euclidean distances for spatial proximity, linking points to neighbors within a set radius, and incorporating cosine similarity in feature space. Weighted by spatial distances, the resulting $\mathcal{W}$ robustly combines spatial and semantic information for clustering.

\noindent \textbf{MinCut clustering} uses $\mathcal{W}$ and $\mathcal{F}_R$ from graph $\mathcal{G}$ to create semantically and geometrically coherent clusters, aggregating $\mathcal{F}_R$ into $K$ meaningful groups.

Specifically, we perform message passing to propagate information across connected nodes to refine $\mathcal{F}_R$, leveraging the geometric and semantic relationships in $\mathcal{W}$ by:
\begin{equation}
    \text{out} = \text{ReLU}(\mathcal{W} \cdot \mathbf{W}_m \mathcal{F}_R) + \mathbf{W}_s \mathcal{F}_R,
\end{equation}
where $\mathbf{W}_m$ and $\mathbf{W}_s$ are learnable weight matrices for mixing and skip connections. The output $\text{out}$ refines the characteristics of each node by integrating information from its local neighborhood, effectively capturing both semantic and spatial relationships.
To establish clusters, we further transform this refined feature set to produce a score matrix $S {\in} \mathbb{R}^{N {\times} K}$ that assigns each node to one of \(K\) clusters:
\begin{equation}
    S = \text{sofmax}\left({\mathbf{W}_2(\text{ReLU}(\mathbf{W}_1 \cdot \text{out}))}/{\tau}\right),
\end{equation}
where \(\mathbf{W}_1\) and \(\mathbf{W}_2\) are learnable projection weights that control the transformation into cluster space, and \(\tau\) is a temperature parameter that tunes the sharpness of the assignment distribution.
Using the assignment scores in \(S\), we then calculate the pooled center points \(\bar{\mathcal{C}}^x = \{\bar{\mathcal{C}}^x_j | j = 1, 2, \dots, K\}, x \in \{a, b\}\), which represent the centroids of each cluster and provide a consolidated spatial representation of the clusters. Specifically, we aggregate the original center points into cluster centroids as follows:
\begin{equation} \label{eq:center}
    \bar{\mathcal{C}}^x_j = \frac{1}{\sum_{i=1}^{N} S^x_{ij}} \sum_{i=1}^{N} S^{\top}_{ij} \mathcal{C}_i, \quad \bar{\mathcal{C}}^x_j \in \mathbb{R}^{K \times 3}.
\end{equation}
This step aggregates information from multiple nodes into compact representative points, where each row of $\bar{\mathcal{C}}$ corresponds to the center of a distinct cluster. Applying MinCut clustering on the geo-semantic graph produces a robust structure that mitigates semantic ambiguities while preserving geometric distinctions. This integration is beneficial for complex shapes in 3D scenes, resulting in a feature space that accurately reflects the data’s intrinsic structure.

\myparagraph{Cluster center reconstruction}
Unlike conventional MAE methods focused on reconstructing individual points, our approach focuses on the reconstruction of cluster centers (\ie, cluster centroids \(\bar{\mathcal{C}}\)), based on the assumption that these centers should form a balanced and uniform partition of the point cloud. Focusing on centroids yields a more structured representation of the underlying geometry.
The reconstruction objective for cluster centers is defined as:
\begin{equation}
    \mathcal{L}_{\text{cts}} = \text{chamfer}(\mathcal{C}^a, \bar{\mathcal{C}}^b) + \text{chamfer}(\mathcal{C}^b, \bar{\mathcal{C}}^a),
\end{equation}
where \(\text{chamfer}(\cdot)\) denotes the Chamfer distance (see \textit{Supp. Mat.} for details). This objective encourages consistent cluster center reconstruction across different masked versions of the same point cloud, reinforcing both geometric and semantic alignment.
Reconstructing cluster centers instead of individual points, this method achieves a compact, high-level view of the data structure that aligns with the clustering objectives outlined in previous sections.

\myparagraph{Cluster assignment reconstruction}
To compute the final assignment of the cluster, we first derive the cluster assignment matrix $\Gamma$, which reflects the probability that each node will be assigned to each cluster, balancing the semantic and geometric information captured through the previous graph construction and message passing steps.
Assuming that each point cloud is divided into approximately equal-sized partitions of \(\left\lfloor \frac{N}{K} \right\rfloor\) elements, we minimize the distance between each point \(\mathcal{C}^x_i\) and its assigned cluster centroid \({\bar{\mathcal{C}}}^y_j\), subject to balanced assignments across clusters, \ie:
\begin{equation}
\label{eq:assignment}
\begin{aligned}
    & \min_{\Gamma} \frac{1}{N}\sum_{i=1}^{N}\sum_{j=1}^{K}\|{\mathcal{C}}^x_i - \bar{\mathcal{C}}^y_j\|^2_2 \Gamma^{xy}_{ij}, \\
    & \text{s.t.} \quad {{\Gamma}^{xy}}^\top \bm{1}_N = \frac{1}{K} \bm{1}_K, \quad {\Gamma} \bm{1}^{xy}_K = \frac{1}{N} \bm{1}_N,
\end{aligned}
\end{equation}
where \( x {=} a, y {=} b \) or \( x {=} b, y {=} a \). This constraint enforces equal distribution of points across clusters, aligning each node with its most suitable cluster centroid. 
To efficiently solve this optimization problem, we utilize an enhanced version of the Sinkhorn-Knopp algorithm~\cite{cuturi2013sinkhorn}.

To ensure consistency between bidirectional assignments, we define the following loss function:
\vspace{-2mm}
\begin{equation}
    \mathcal{L}_{\text{ass}} {=} {-}\frac{1}{N} \sum^{N,K}_{i,j=1} \text{stop}({\Gamma}^{ab}_{ij}) \log s^{ab}_{ij} {+} \text{stop}({\Gamma}^{ba}_{ij}) \log s^{ba}_{ij},
    \vspace{-2mm}
\end{equation}
where \(\text{stop}(\cdot)\) halts gradient propagation during backpropagation for stability. By minimizing \(\mathcal{L}_{\text{ass}}\), the assignment of the cluster is refined and reconstructed, ensuring the alignment between the geometric and semantic spaces.

\begin{algorithm}[t]
\caption{Contrastive PyTorch-like Pseudocode}
\label{alg:code}
\vspace{-1mm}
\begin{lstlisting}[language=python]
# (*@$f_\varphi$@*): backbone
# (*@$h$@*)  : max-pooling operation
for x in loader:  # load a batch x with n samples
    (*@$\mathcal{T}^{a}$@*), (*@$\mathcal{T}^{b}$@*) = aug((*@$\mathcal{T}$@*)), aug((*@$\mathcal{T}$@*))  # random mask aug
    [(*@$f^{a}_{cls}$@*), (*@$\mathcal{F}^{a}$@*)], [(*@$f^{b}_{cls}$@*), (*@$\mathcal{F}^{b}$@*)] = (*@$f_\varphi$@*)((*@$\mathcal{T}^{a}$@*)), (*@$f_\varphi$@*)((*@$\mathcal{T}^{b}$@*)) 
    (*@$z^{a}$@*), (*@$z^{b}$@*) = pooling((*@$\mathcal{F}^{a}$@*)), pooling((*@$\mathcal{F}^{b}$@*)) 
    L = D((*@$f^{a}_{cls}$@*), (*@$z^{b}$@*)) + D((*@$f^{b}_{cls}$@*), (*@$z^{a}$@*))   # loss
    L.backward()  # back-propagate
    update((*@$f_\varphi$@*))  # SGD update
    
def D(f, z):  # negative cosine similarity
    z   = z.detach()  # stop gradient
    s1= cos(f, z)
    f   = f.detach()
    z   = z.requires_grad_() # re-enable gradient 
    s2= cos(f, z) 
    return (2 - s1 - s2).sum(dim=1).mean()
\end{lstlisting}
\vspace{-0.5mm}
\end{algorithm}

\subsection{Global Contrastive Constraint}
\label{subsec:contrastive}
We devise the proposed global unsupervised learning based on SimSiam~\cite{chen2020exploring}. As shown in Fig.~\ref{fig:framework}, our architecture processes two randomly masked views, ${\mathcal{T}}^a$ and ${\mathcal{T}}^b$, derived from a single set of point patch embeddings ${\mathcal{T}}$.
The contrastive learning paradigm in \ourmethod{} is outlined in Alg.~\ref{alg:code}, where $\mathcal{L}_{\text{contras}}$ denotes the contrastive loss.

\subsection{Optimization Objectives}
\label{subsec:opjectives}
By considering both the masked clustering and contrastive learning, the standard ViTs encoder can be pretrained by minimizing the following objectives:
\begin{equation}
    \mathcal{L}_{total} = \mathcal{L}_{ass} +  \mathcal{L}_{cts} + \mathcal{L}_{contras}.
\end{equation}

%% file: secs/4_experiments.tex
\section{Experiments}
\label{sec:exp}
In this section, we first validated \ourmethod{} on 4 downstream tasks, \ie, 
3D part segmentation, 
semantic segmentation, 
object detection, and 
object classification (Sec.~\ref{sec:downstream_tasks}). Then we present the ablation analysis (Sec.~\ref{subsubsec:ablations}). 
The implementation and other details are provided in our \textit{Supplementary Materials (Supp. Mat.)}.

\subsection{Downstream tasks}\label{sec:downstream_tasks}
We assess the efficacy of our method by fine-tuning our pre-trained models on several downstream tasks, \ie,
part segmentation, 
semantic segmentation, 
classification, and few-shot learning.
We compare \ourmethod{} with existing methods with 
Point Trans.~\cite{engel2021point}, 
PCT~\cite{sauder2019self}, 
PointVit-OcCo~\cite{wang2021unsupervised}, 
Point-BERT~\cite{yu2022point}, 
MaskPoint~\cite{liu2022masked}, 
Point-MAE~\cite{pang2022masked}, 
Point-M2AE~\cite{zhang2022point}, 
Point-MA2E~\cite{MA2E}, 
Point-CMAE~\cite{ren2024bringing},
OcCo~\cite{wang2021unsupervised}, and CrossPoint~\cite{afham2022crosspoint}.

\begin{table}[!t]
    \small
    \centering
    \tabcolsep 7pt
    \caption{Segmentation results on ShapeNetPart and S3DIS Area 5.  mIoU\textsubscript{C} (\%) and mIoU\textsubscript{I} (\%) for \textbf{Part Segmentation}, and  mAcc (\%) and mIoU (\%) for \textbf{Semantic Segmentation}.}
    \vspace{-1mm}
    \label{tab:segmentation}
    \scalebox{0.99}{
    \begin{tabular}{lcccc}
    \toprule
    \textbf{Methods} & \multicolumn{2}{c}{{Part Seg.}} & \multicolumn{2}{c}{Semantic Seg.} \\
    & mIoU\textsubscript{C} & mIoU\textsubscript{I} & mAcc & mIoU \\
    \midrule
    \multicolumn{5}{c}{\textit{with single-modal self-Supervised Learning}} \\
    \midrule
    Scratch & 83.4  & 84.7  & 68.6  & 60.0  \\
    Point-BERT \cite{yu2022point}       & 84.1  & 85.6  & -     & -     \\
    MaskPoint \cite{liu2022masked}       & 84.4  & 86.0  & -     & -     \\
    Point-MAE \cite{pang2022masked}         & 84.2  & 86.1  & 69.9  & 60.8  \\
    Point-MA2E~\cite{MA2E} & - & 86.4 & - & - \\
    PointGPT \cite{chen2024pointgpt}       & 84.1  & 86.2  & -     & -     \\
    Point-FEMAE \cite{zha2024towards}        & 84.9  & 86.3  & -     & -     \\
    SoftClu~\cite{mei2022data} & - & 86.1 & - & 61.6 \\
    MaskSurf~\cite{zhang2022masksurf} & 84.6 & 86.1 & 69.9 & 61.6 \\
    Point-CMAE~\cite{ren2024bringing} & 84.9 & 86.0 & - & - \\
    PCP-MAE~\cite{zhang2024pcp} & 84.9 & 86.1 & 71.0 & 61.3 \\
    \rowcolor{linecolor}\ourmethod{~} (Ours) & \bf85.2 & \bf86.4 & \bf71.3 & \bf62.1 \\
    \midrule
    \multicolumn{5}{c}{\textit{with hierarchical/multimodal/self-supervised learning}} \\
    \midrule
    Point-M2AE~\cite{zhang2022point} & 84.8 & 86.5 & - & -\\
    CrossPoint \cite{afham2022crosspoint}   & -     & 85.5  & -     & -     \\
    CluRender \cite{mei2024unsupervised}     & -  & 86.9  & -  & 62.6  \\
    PointGPT-L~\cite{chen2024pointgpt} & 84.8 & 86.6 & - & -\\
    ACT~\cite{dong2022autoencoders} & 84.7 & 86.1 & - & -\\
    ReCon~\cite{qi2023contrast}             & 84.8  & 86.4  & 71.1  & 61.2     \\
    PointClustering~\cite{long2023pointclustering} & - & 86.7 & - & 65.6 \\
    \rowcolor{linecolor}\ourmethod{~} (Ours) & \bf85.6 & \bf87.4 & \bf72.2 & \bf66.4 \\
    \bottomrule
    \end{tabular}
    }
\end{table}

\begin{figure*}[t]
    \centering
    \tabcolsep 4pt
    \renewcommand{\arraystretch}{0.5}
    \vspace{-1mm}
    \begin{tabular}{@{}c@{}c@{}c@{}c@{}c@{}c@{}c@{}c@{}}
    
    \begin{overpic}[width=.125\textwidth]{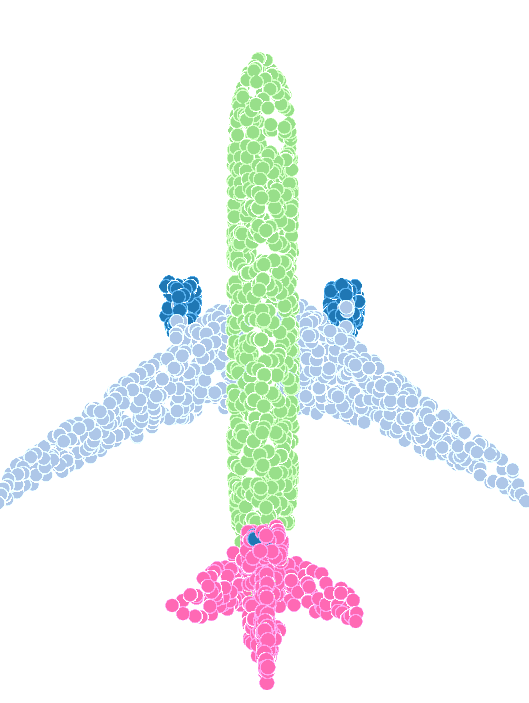}
        \put(0,40){\rotatebox{90}{\color{black}\footnotesize \textbf{GT}}}
    \end{overpic} &
    \begin{overpic}[width=.125\textwidth]{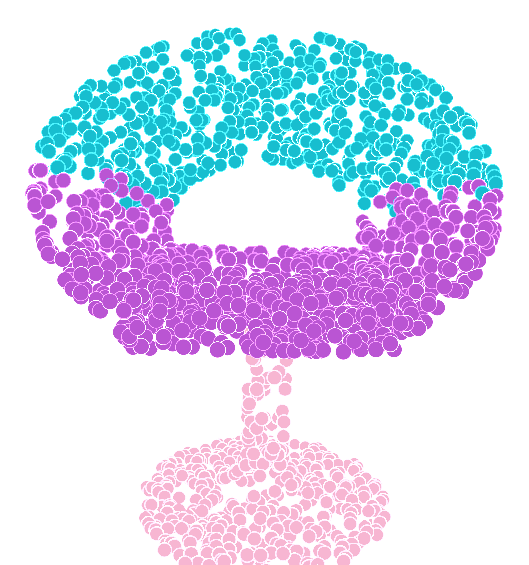}
    \end{overpic} &
    \begin{overpic}[width=.125\textwidth]{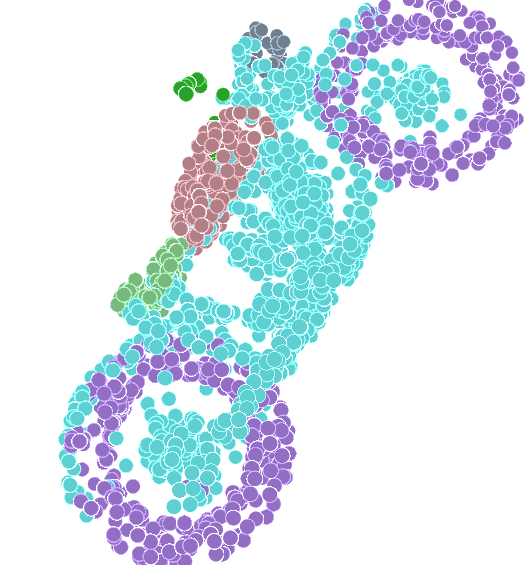}
    \end{overpic} &
    \begin{overpic}[width=.125\textwidth]{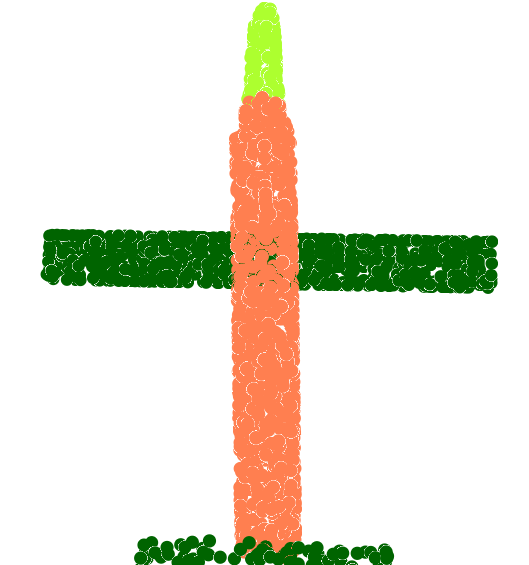}
    \end{overpic} &
    \begin{overpic}[width=.125\textwidth]{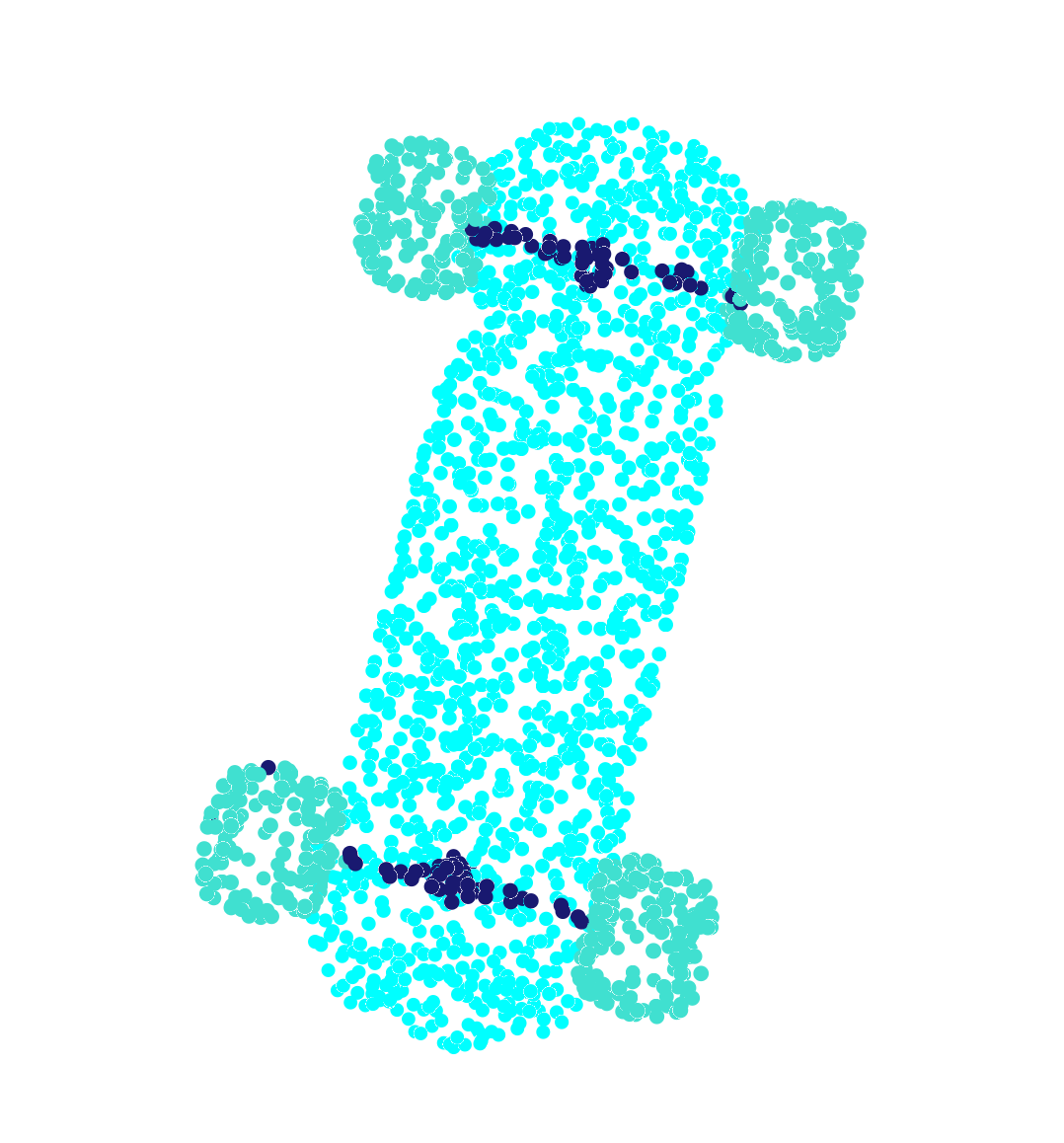}
    \end{overpic} &
    \begin{overpic}[width=.125\textwidth]{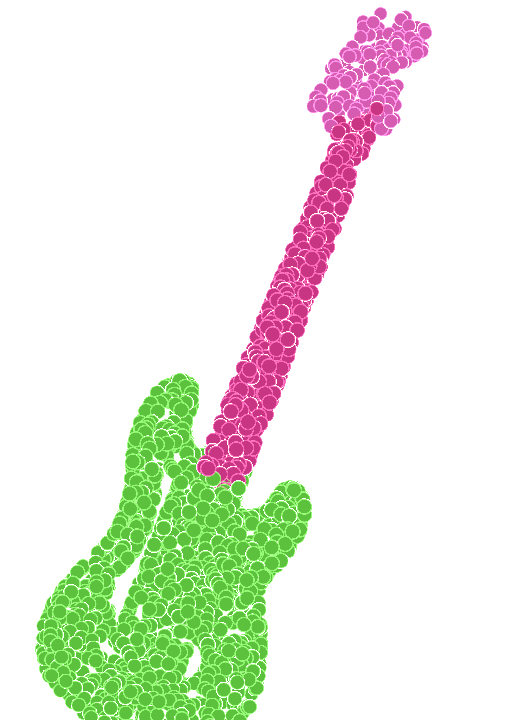}
    \end{overpic} &
    \begin{overpic}[width=.125\textwidth]{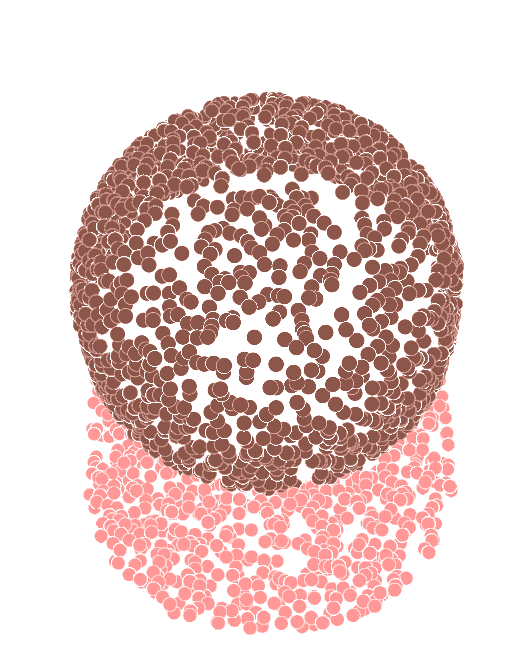}
    \end{overpic} &
    \begin{overpic}[width=.125\textwidth]{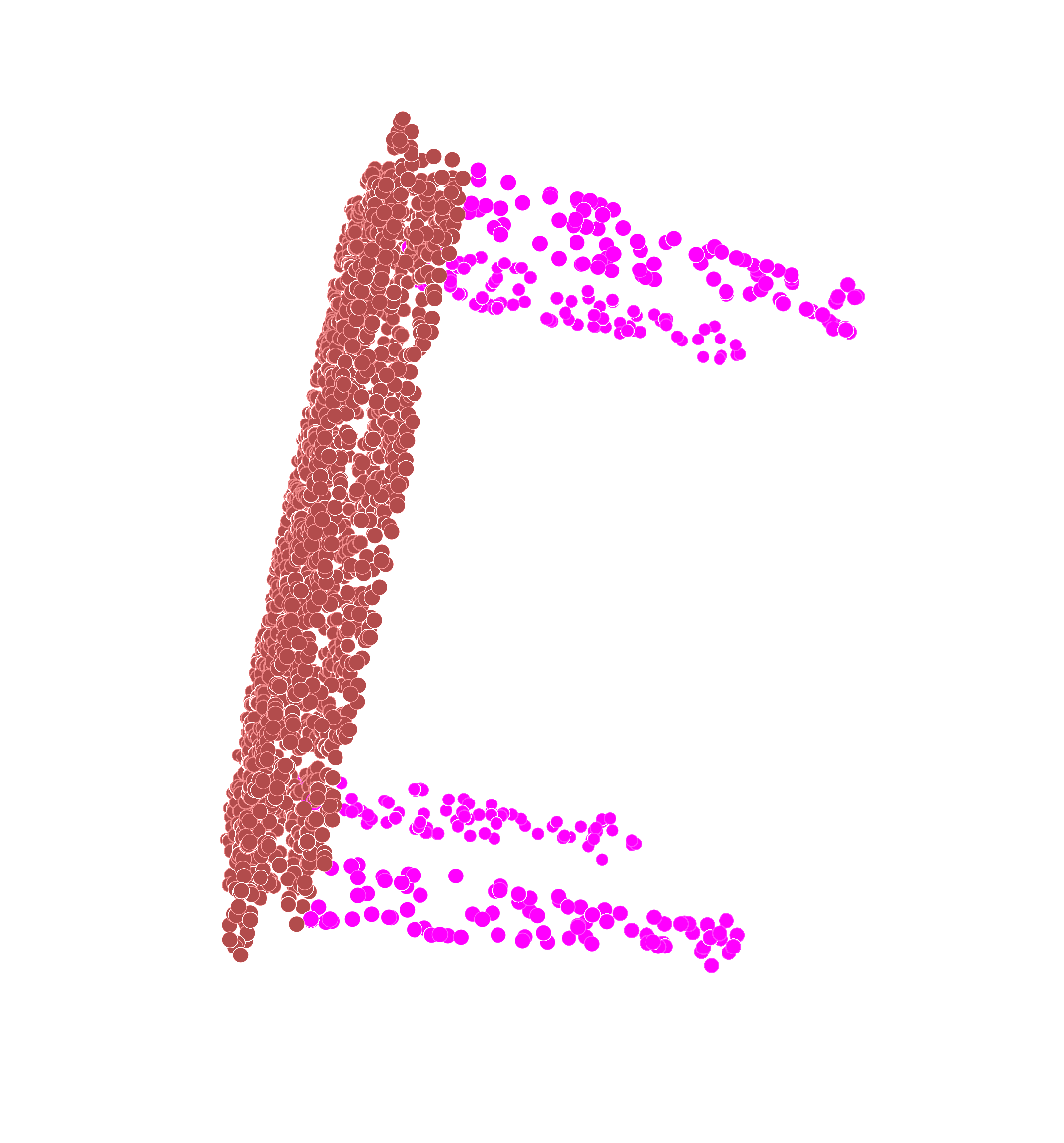}
    \end{overpic} \\[-4mm]

    \begin{overpic}[width=.125\textwidth]{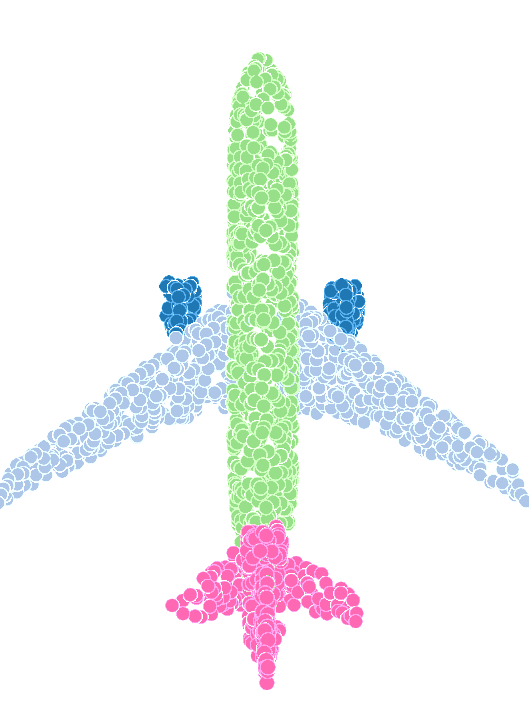}
        \put(0,40){\rotatebox{90}{\color{black}\footnotesize \textbf{PointMAE}}}
    \end{overpic} &
    \begin{overpic}[width=.125\textwidth]{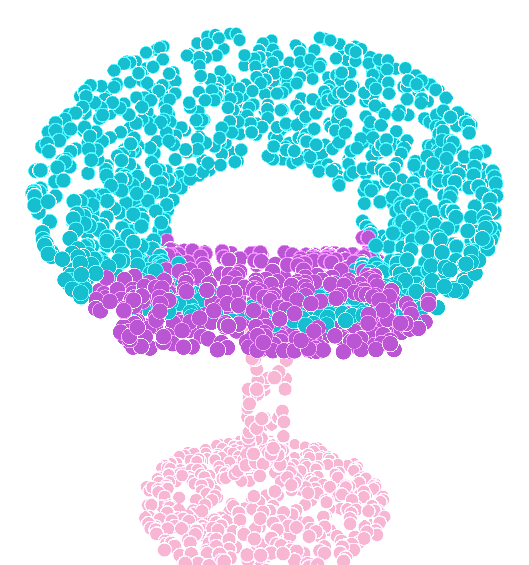}
        \put(18,38){\color{blue}\dashbox{30}(60,20){}}  
    \end{overpic} &
    \begin{overpic}[width=.125\textwidth]{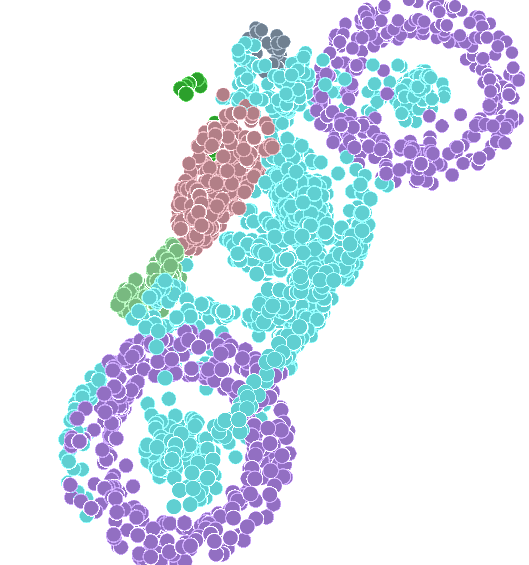}
    \end{overpic} &
    \begin{overpic}[width=.125\textwidth]{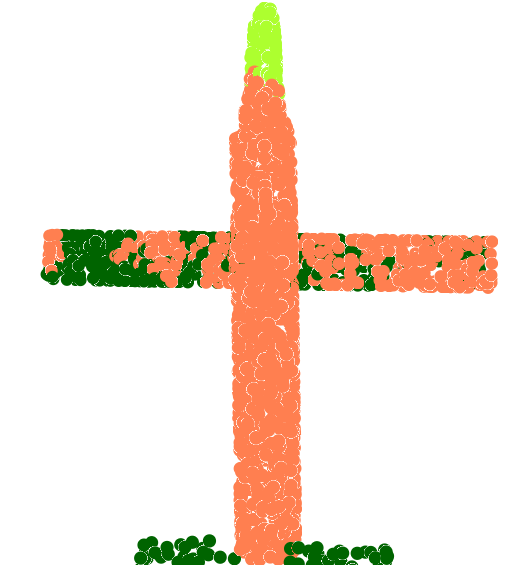}
        \put(8,45){\color{blue}\dashbox{30}(80,20){}}  
    \end{overpic} &
    \begin{overpic}[width=.125\textwidth]{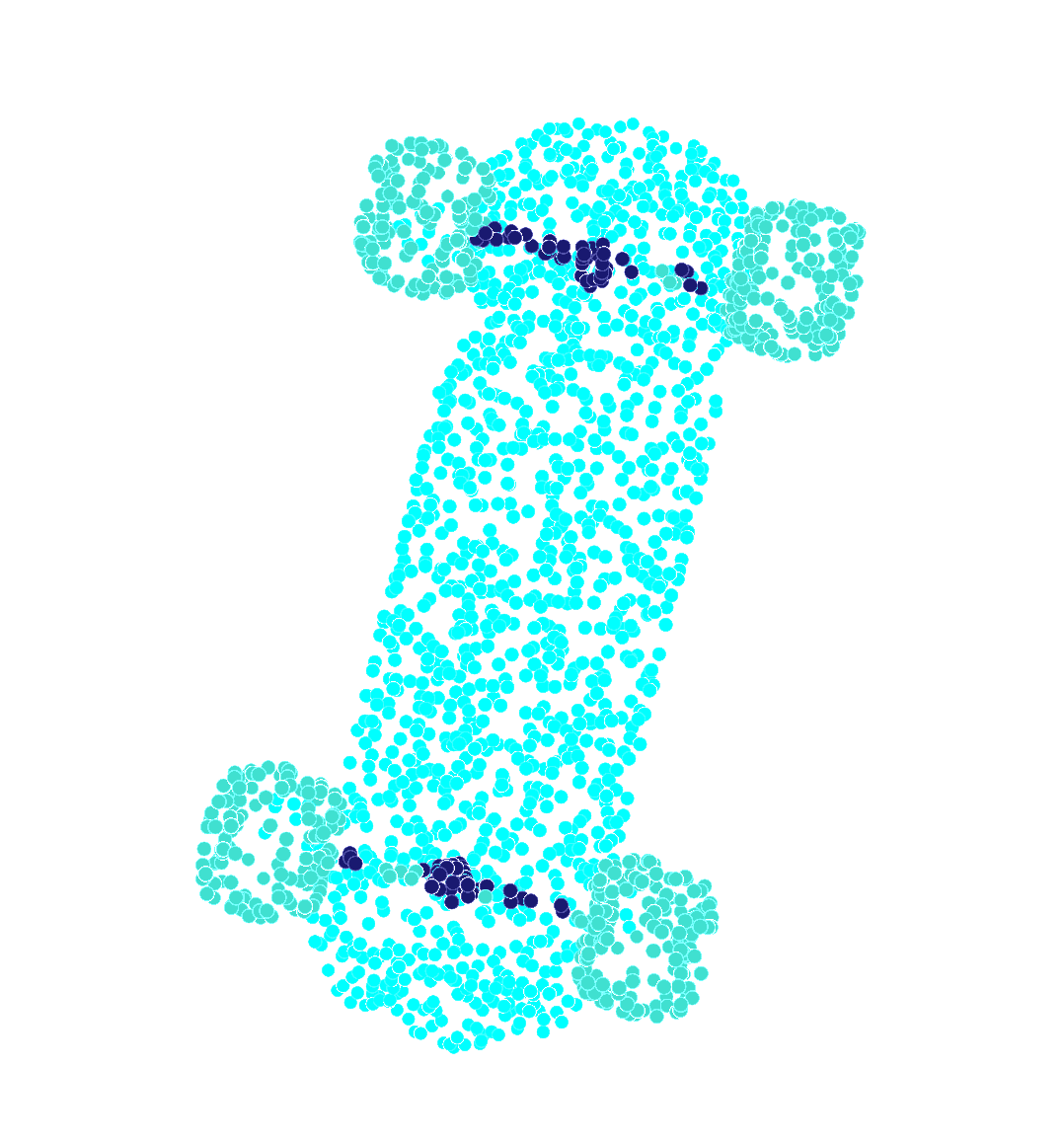}
    \end{overpic} &
    \begin{overpic}[width=.125\textwidth]{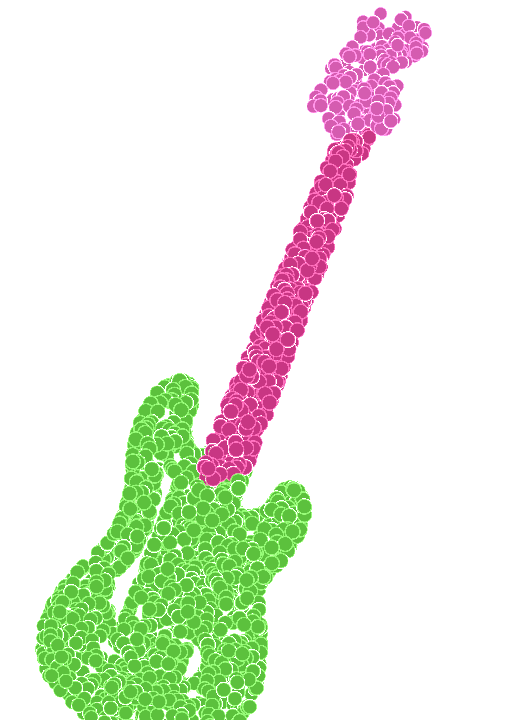}
    \end{overpic} &
    \begin{overpic}[width=.125\textwidth]{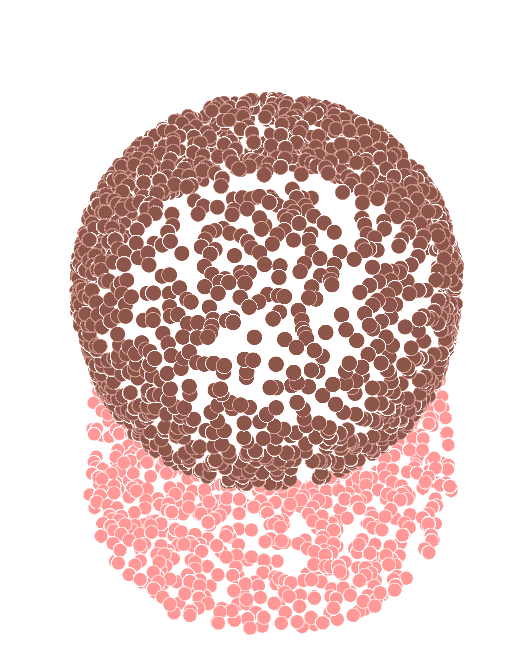}
    \end{overpic} &
    \begin{overpic}[width=.125\textwidth]{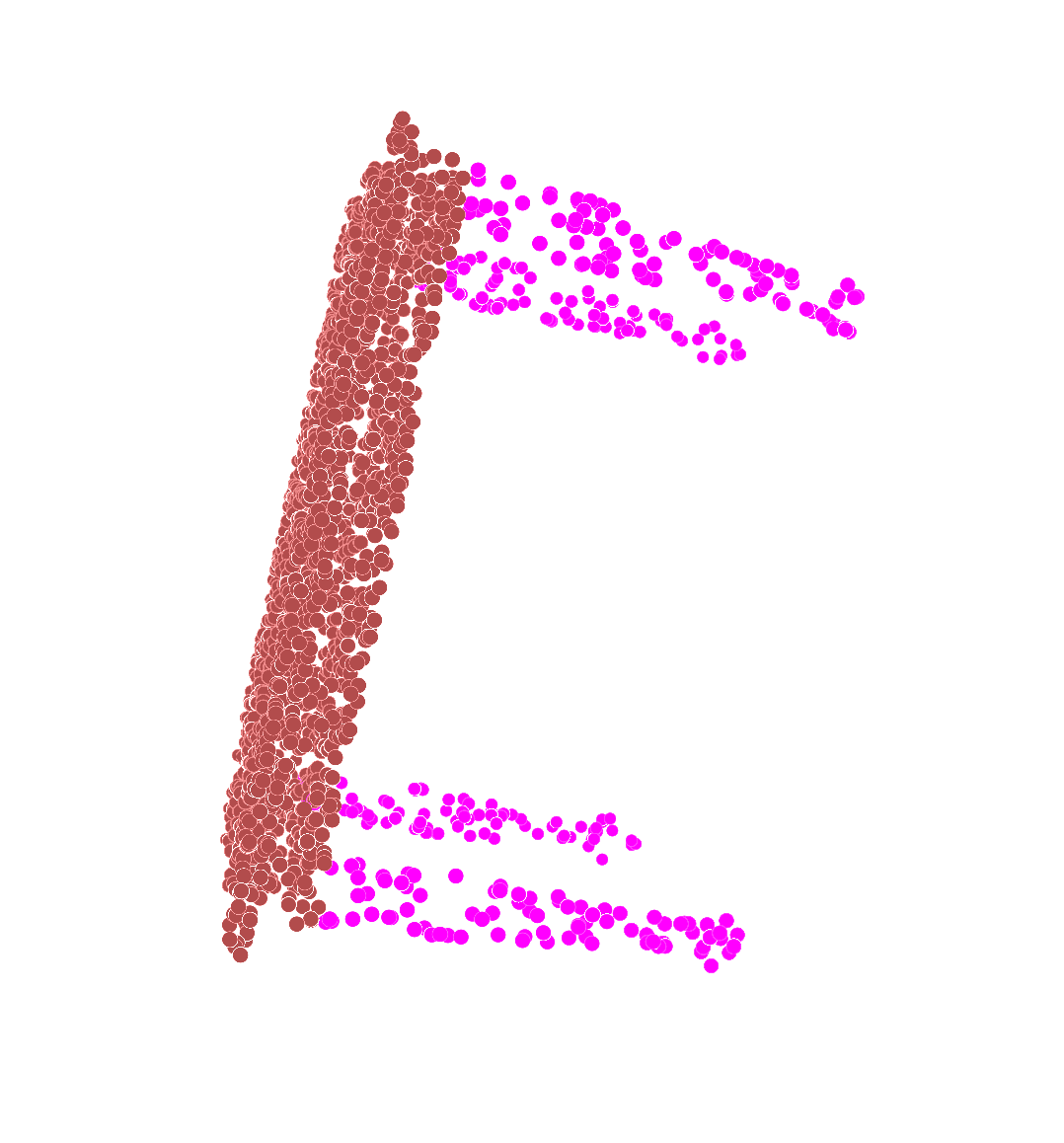}
    \end{overpic} \\[-4mm]

    \begin{overpic}[width=.125\textwidth]{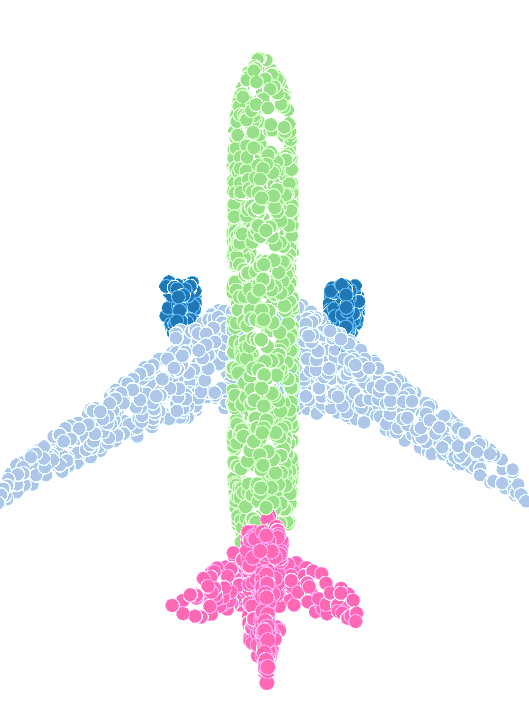}
        \put(0,40){\rotatebox{90}{\color{black}\footnotesize \textbf{\ourmethod}}}
    \end{overpic} &
    \begin{overpic}[width=.125\textwidth]{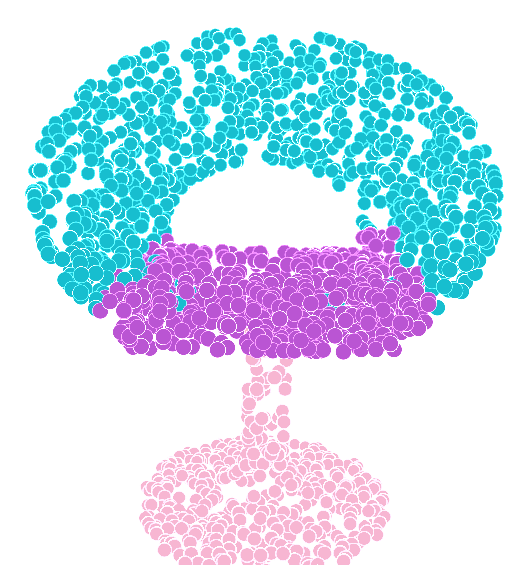}
    \end{overpic} &
    \begin{overpic}[width=.125\textwidth]{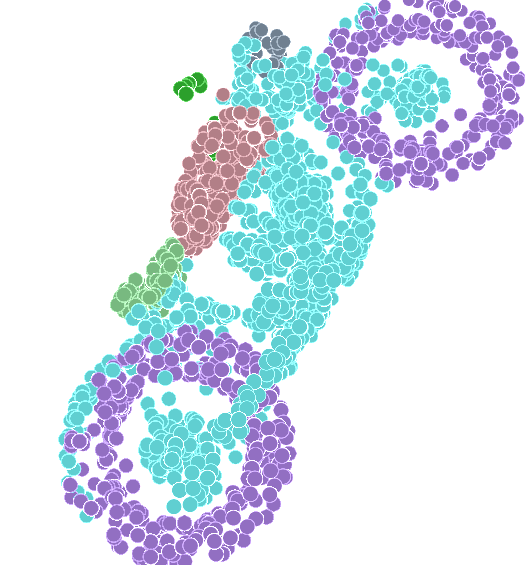}
    \end{overpic} &
    \begin{overpic}[width=.125\textwidth]{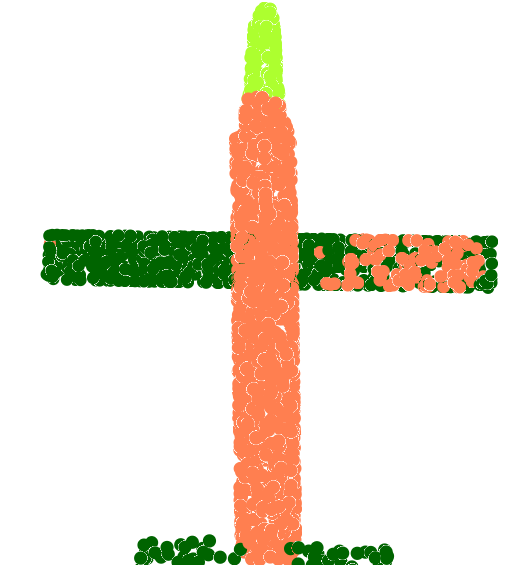}
        \put(50,45){\color{blue}\dashbox{30}(40,20){}}  
    \end{overpic} &
    \begin{overpic}[width=.125\textwidth]{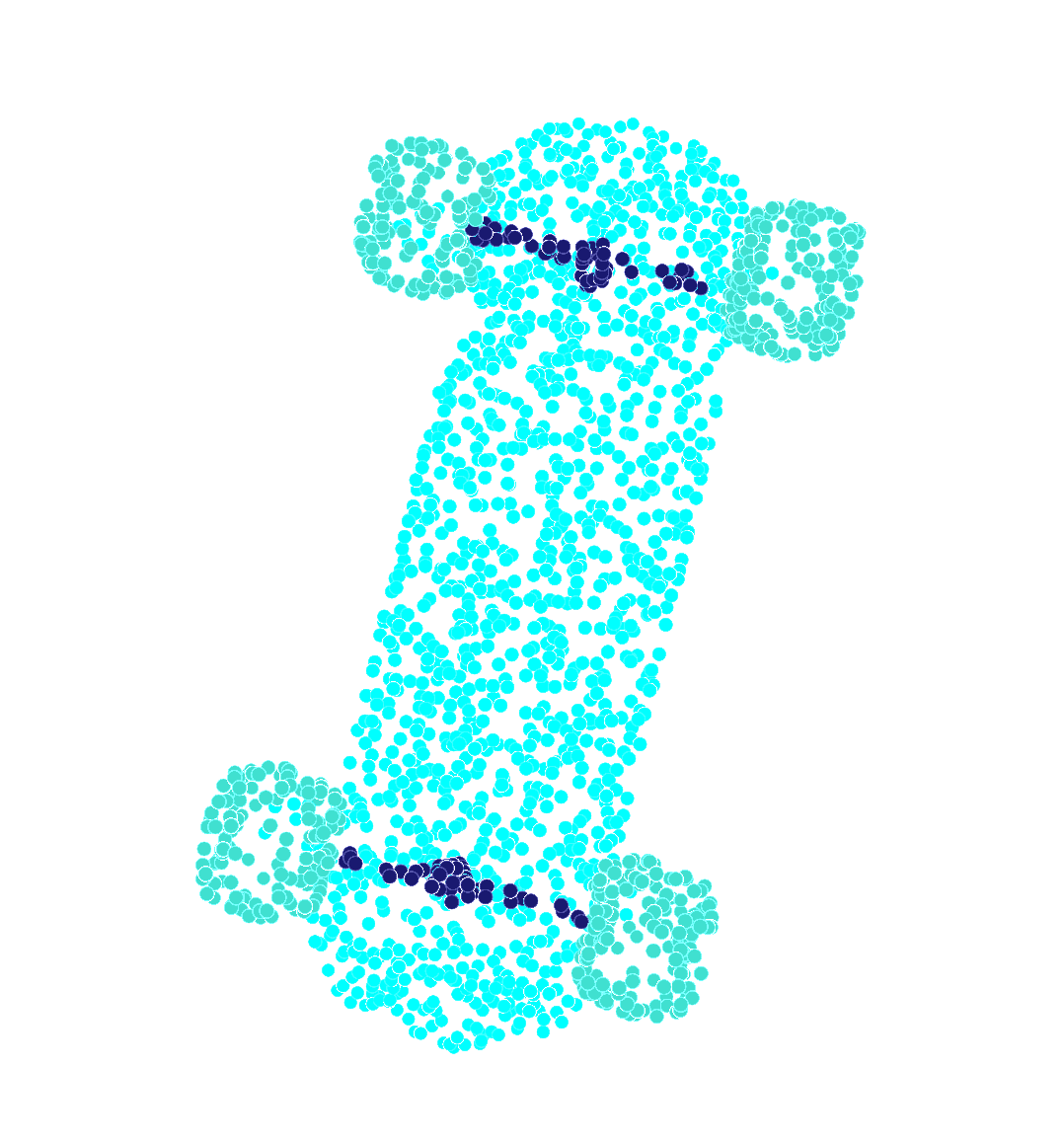}
    \end{overpic} &
    \begin{overpic}[width=.125\textwidth]{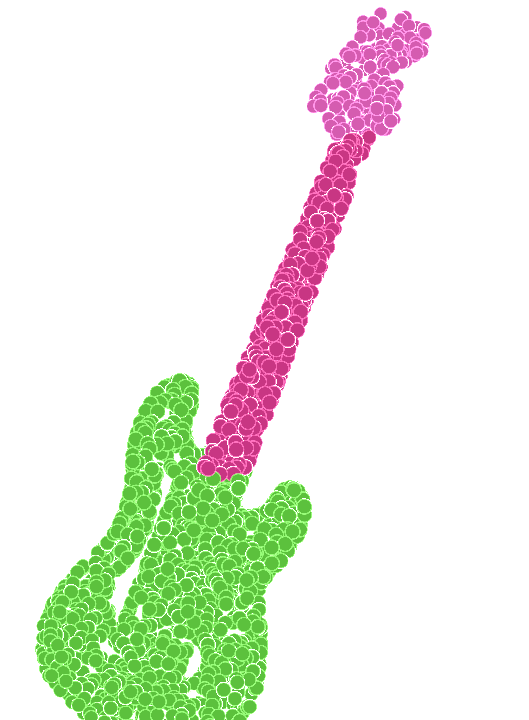}
    \end{overpic} &
    \begin{overpic}[width=.125\textwidth]{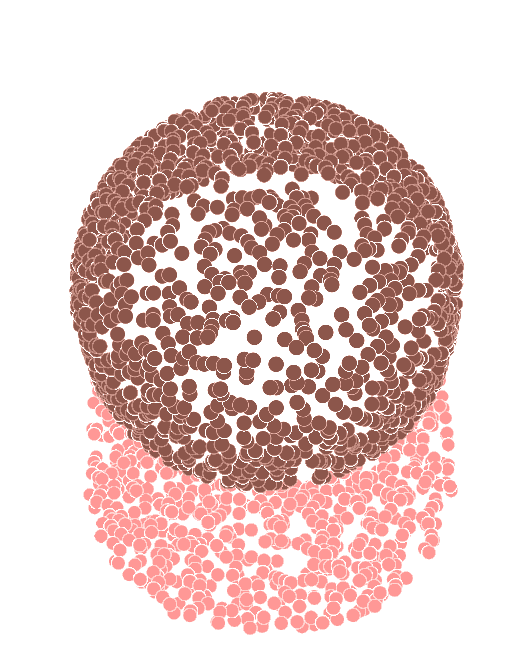}
    \end{overpic} &
    \begin{overpic}[width=.125\textwidth]{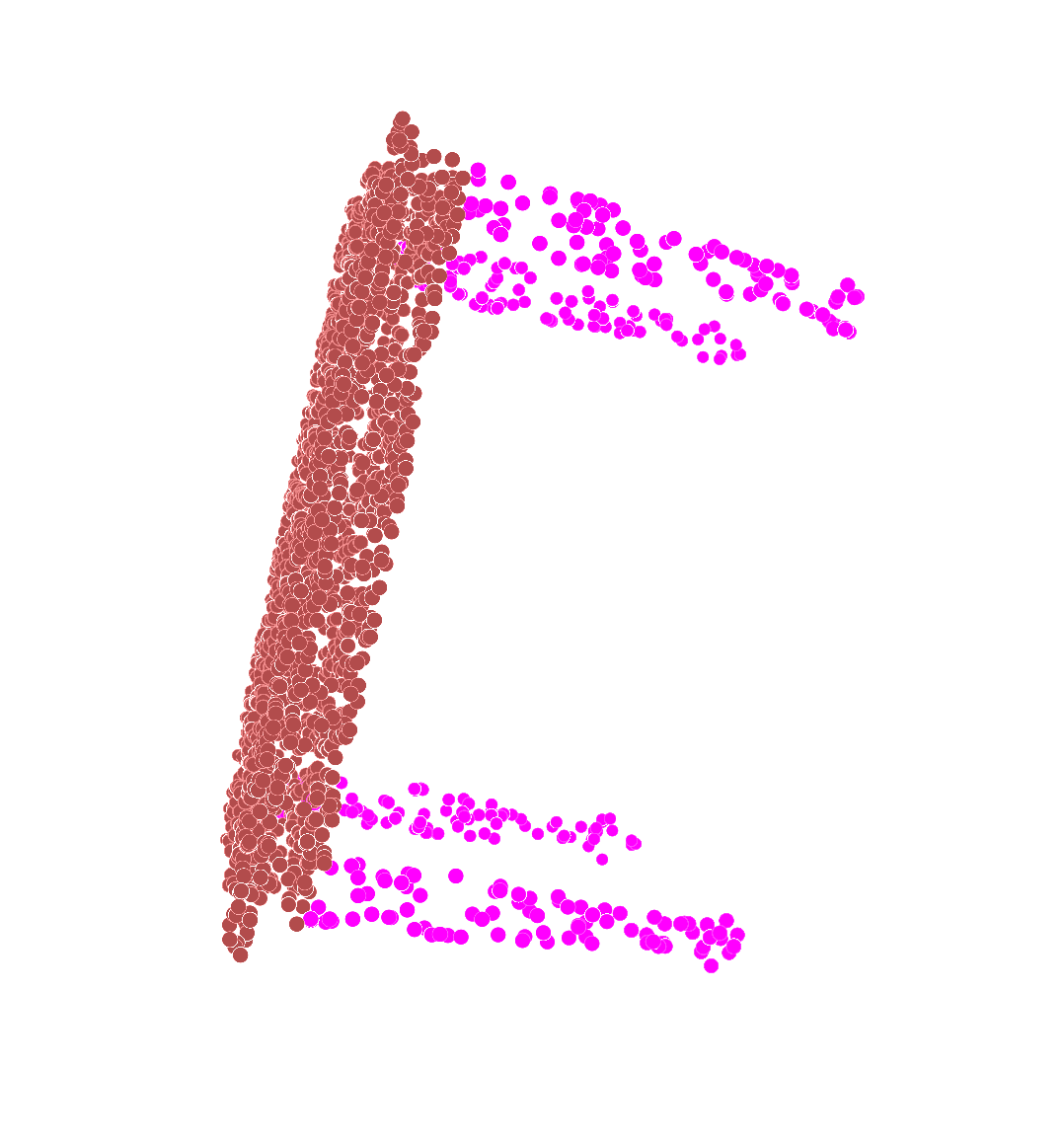}
    \end{overpic} \\
    
    \end{tabular}
    \vspace{0mm}
    \caption{Part segmentation results on ShapeNet~\cite{chang2015shapenet} comparing \ourmethod (bottom row) using the ViT encoder with PointMAE and ground-truth annotations (top row). Blue dashed boxes highlight areas where segmentation failed.}
    \label{fig:pargset}
    \vspace{0mm}
\end{figure*}

\noindent \textbf{Part segmentation.} 
Since one component of \ourmethod{} involves clustering, which tends to capture regional semantic information, we initially assess our pre-trained models on part segmentation tasks.
We evaluate the segmentation performance of the proposed \ourmethod{} on the 
ShapeNetPart dataset~\cite{chang2015shapenet}, 
which contains 16,881 objects, each represented by 2,048 points across 16 categories, with a total of 50 parts. 
The segmentation head used in our method follows the design of Point-MAE, remaining relatively simple without employing any propagating operations~\cite{qi2017pointnet++} or DGCNN~\cite{wang2019dynamic}. 
Following PointMA2E~\cite{pang2022masked}, we extract features from the 4th, 8th, and 12th layers of the Transformer block and concatenate these multi-level features to form the final representation.
The mean intersection over union across all categories, \ie, mIoU\textsubscript{C} (\%), and the mean IoU across all instances, \ie, mIoU\textsubscript{I} (\%), are reported.
Tab.~\ref{tab:segmentation} \textbf{(Part Seg.}) presents the part segmentation results of \ourmethod{} compared to alternative approaches on ShapeNetPart. \ourmethod{} outperforms all other methods in terms of mIoU with a standard ViTs architecture.
With \ourmethod, we achieve the highest mIoU\textsubscript{I} of 87.4\%, surpassing the second best method (CluRender) by \textbf{0.5\%}. Furthermore, \ourmethod{} reaches 85.6\% in mIoU\textsubscript{C}, outperforming Point-FEMAE and other competitive baselines. These results highlight the superior performance of \ourmethod{} in part segmentation.
The qualitative results in Fig.~\ref{fig:pargset} indicate that compared to GT or PointMAE, our method produced consistent predictions across various object categories, especially for complex shapes like chairs or motorcycls.

\noindent \textbf{Semantic segmentation} involves classifying each point in 3D scenes according to its semantic category. Specifically, we assess \ourmethod{} via using the S3DIS benchmark dataset~\cite{armeni_cvpr16}.
which contains 3D scans of six indoor areas captured with a Matterport scanner, totaling 271 rooms across 13 semantic classes.
Following the pre-processing, post-processing, and training protocols in \cite{wang2021unsupervised}, we divide each room into $1 \, \text{m} \times 1 \, \text{m}$ blocks and use 4,096 points as input to the model. 
We fine-tune the pre-trained model on Areas [1, 2, 3, 4, 6], and evaluate it on Area 5.
Tab.~\ref{tab:segmentation} (\textbf{Semantic Seg.}) presents the semantic segmentation results of \ourmethod{} compared with other methods on Area 5 of the S3DIS dataset. We report the overall accuracy (OA) and mean class intersection over union (mIoU).
\ourmethod{} outperforms all single-modal self-supervised learning approaches, achieving the best mAcc of 72.2\% and mIoU of 66.3\%.
Specifically, \ourmethod{} surpasses the second-best method, PointClustering~\cite{long2023pointclustering}, by \textbf{0.9\%} in mIoU and ReCon~\cite{qi2023contrast}, by \textbf{1.1\%} in mAcc, demonstrating its superiority in semantic segmentation. 
These results demonstrate the robust generalizability and consistent accuracy of \ourmethod{} in handling complex, diverse scenes—even without multi-modal input.

\begin{table}[!t]
    \centering
    \small
    \caption{3D \textbf{object detection} Results on ScanNet validation set.}\label{tab:detection}
    \vspace{-1mm}
    \scalebox{0.99}{
    \begin{tabular}{lccc}
    \toprule
    \textbf{Methods} & \textbf{Pre Dataset} & \textbf{AP25} & \textbf{AP50} \\
    \midrule
    Scratch & - & 62.1 & 37.9 \\
    Point-BERT~\cite{PointBERT} & ScanNet & 61.0 & 38.3 \\
    MaskPoint~\cite{liu2022masked} & ScanNet & 63.4 & 40.6 \\
    \rowcolor{linecolor}\ourmethod{~} (Ours) & ShapeNet & \bf63.9 & \bf41.4 \\
    \midrule
    MaskPoint (12 Enc)~\cite{liu2022masked} & ScanNet & 64.2 & 42.1 \\
    \rowcolor{linecolor}\ourmethod{~} (12 Enc) & ShapeNet & \bf65.0 & \bf43.2 \\
    \bottomrule
    \end{tabular}
    }
    \vspace{0mm}
\end{table}

\noindent \textbf{3D object detection.} We adopt 3DETR~\cite{misra2021end} as our baseline, and follow MaskPoint~\cite{liu2022masked} by replacing the 3DETR encoder with our pre-trained encoder, then fine-tuning on ScanNetV2~\cite{dai2017scannet} for a fair comparison.
Unlike MaskPoint and Point-BERT, which are pre-trained on the ScanNet-Medium dataset (within the same domain as ScanNetV2), our method is pre-trained only on ShapeNet and fine-tuned on the ScanNetV2 training set.
As shown in Tab.~\ref{tab:detection}, \ourmethod{} achieves an AP50 of 41.4, outperforming Point-BERT by \textbf{3.1\%} and MaskPoint by \textbf{0.8\%} in AP50. Additionally, with a 12-layer encoder, \ourmethod{} reaches an AP50 of 43.2, which represents an improvement of \textbf{1.1\%} over MaskPoint’s AP50 of 42.1. These results shows that \ourmethod{} provides strong transferability for 3D scene understanding tasks.

\input{tabs/scanobjectnn}
\noindent \textbf{Object Classification on Real Dataset.}
We assess \ourmethod{} on the ScanObjectNN dataset~\cite{uy2019revisiting}, 
which includes approximately 15,000 real-world objects in 15 categories. 
which consists of 3 subsets: OBJBG (objects with background), OBJONLY (objects without background), and PBT50RS (objects with background and artificially applied perturbations). 
To ensure comparison fairness, we report the results without a voting strategy~\cite{liu2019relation}, and each input point cloud is sampled to 2048 points.
Tab.~\ref{tab:scanobjectnn} shows that \ourmethod{} achieves competitive performance:
\textit{1) With Full Tuning}: \ourmethod{} outperforms nearly all baseline methods across all subsets, achieving 90.88\% on OBJBG, 90.36\% on OBJONLY, and 87.12\% on PBT50RS without data augmentation, surpassing Point-MAE and Point-CMAE in these settings. Compared to MaskPoint, \ourmethod{} shows a significant improvement, especially on PBT50RS, with a \textbf{1.17\%} increase in accuracy. Once fine-tuned with rotation data augmentation (used by ACT~\cite{dong2022autoencoders} and PointGPT~\cite{chen2024pointgpt}), \ourmethod{} achieves state-of-the-art performance of 93.59\% on OBJBG, 92.04\% on OBJONLY, and 89.28\% on PBT50RS, outperforming ACT by \textbf{0.3\%} on OBJBG and by \textbf{1.07\%} on PBT50RS.
\textit{2) With MLP-Linear and MLP-3 Fine-Tuning}: In both MLP-Linear and MLP-3 settings, \ourmethod{} demonstrates strong generalization. It achieves 84.18\% on OBJBG, 84.34\% on OBJONLY, and 76.79\% on PBT50RS in MLP-Linear, and 87.44\% on OBJBG, 86.92\% on OBJONLY, and 79.88\% on PBT50RS with MLP-3. These results highlight the effectiveness of our pre-training, with \ourmethod{} achieving robust performance on ScanObjectNN by capturing both shape and semantic details across diverse objects.
More comparisons, including classification and few-shot learning on the synthetic ModelNet40 dataset~\cite{wu20153d}, are provided in \textit{Supp. Mat}.

\subsection{Ablation study}
\label{subsubsec:ablations}
We perform our ablation study on the challenging PBT50RS subset of ScanObjectNN for hard classification tasks, utilizing standard ViTs in a single-modal setup, as shown in Tab.~\ref{tab:scanobjectnn}. 

\begin{figure}[!t]
    \centering
    \includegraphics[width=1.0\linewidth]{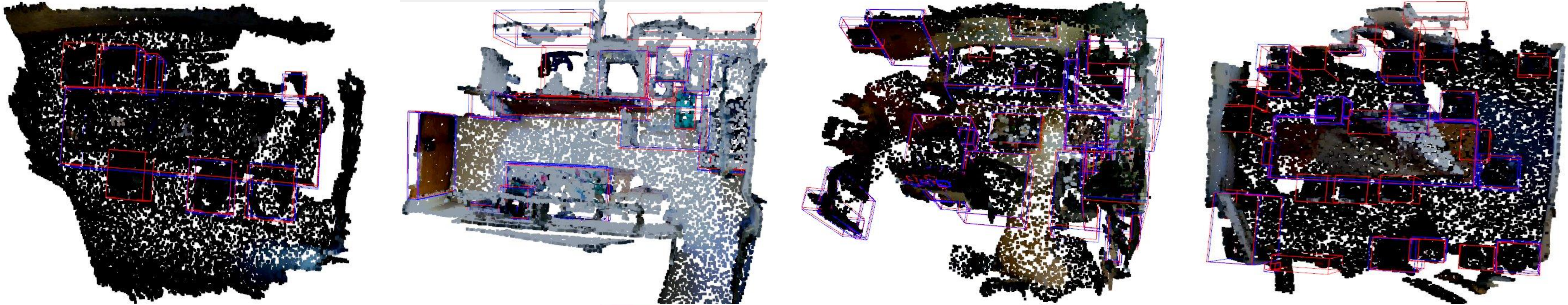}
    \vspace{-1mm}
    \caption{Four detection samples from ScanNet. Red: ground truth; blue: predictions. The model robustly handles complex 3D scenes with diverse structures and object layouts.}
    \label{fig:detection}
    \vspace{0mm}
\end{figure}

\begin{figure}[!t]
    \centering
    \includegraphics[width=1.0\linewidth]{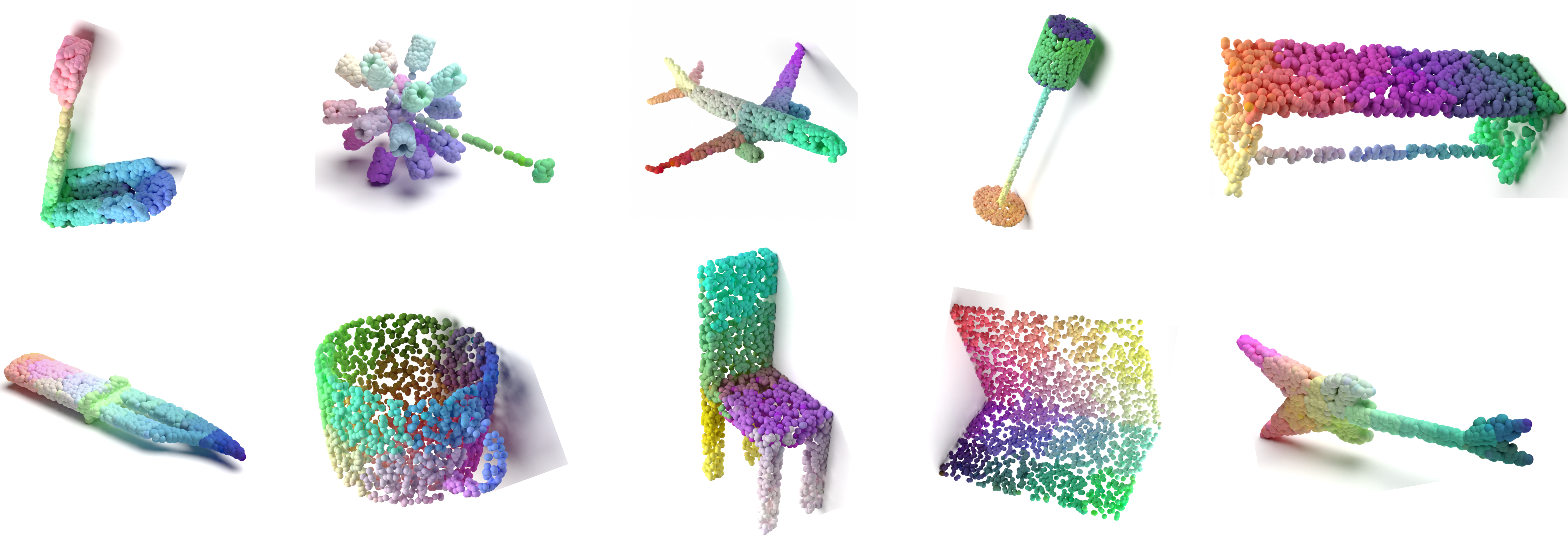}
    \vspace{0mm}
    \caption{Visualization of learned representations. The points are color-coded based on the PCA projection of propagated features from the pre-trained encoder by the proposed \ourmethod{}. Distinct colors correspond to different feature clusters.}
    \label{fig:pca_cluster}
    \vspace{0mm}
\end{figure}

\noindent \textbf{Pretraining results.}
We aim to learn dense features through pretraining without modifying the ViT encoder.
To assess whether our cluster-assignment reconstruction clustering improves the ViT's ability to learn informative features, we visualize the pretrained features. Since the encoder produces patch-level features, we employ the non-learning method Superpoint-Aware Feature Propagation~\cite{mei2025self} to propagate these patch features back to the point level. We then color each point based on the PCA projection of its feature (Fig.~\ref{fig:pca_cluster}). As shown in the figure, the pre-trained encoder organizes points into coherent, semantically meaningful regions (\eg, airplane wings, chair legs, and table surfaces) while preserving geometric boundaries. This demonstrates that the learned representations capture both local geometry and high-level object structure, validating the effectiveness of our pretraining approach.

\myparagraph{Impact of the joint learning.}
To evaluate the effectiveness of our joint learning objective in \ourmethod{}, we examined the individual and combined contributions of global contrast and local clustering. We trained the network under different configurations, such as: 
(\textit{i}) using only global contrast ($\mathcal{L}_{contra}$), 
(\textit{ii}) using only local clustering with 24 clusters ($\mathcal{L}_{ass}$), and 
(\textit{iii}) combining both global contrast and local clustering ($\mathcal{L}_{contra} + \mathcal{L}_{ass} + \mathcal{L}_{cts}$). The results on the PB-T50-RS subset are presented in Tab.~\ref{tab:ab_design}.
Local clustering alone achieved strong performance, underscoring the importance of spatial information in feature learning. However, the joint approach, which integrates global and local clustering, yielded the best performance with an accuracy of 87.12\%, demonstrating the advantage of our combined objective for comprehensive feature representation.

\begin{table}[!t]
    \small
    \centering
    \caption{Ablation study on the impact of joint learning (\ie, with different combinations of the optimization objectives).}
    \label{tab:ab_design}
    \vspace{-1mm}
    \tabcolsep 3pt
    \resizebox{\columnwidth}{!}{%
    \begin{tabular}{c|ccccccc} 
        \toprule
        Loss & $\mathcal{L}_{contra}$ & $\mathcal{L}_{ass}$ & $\mathcal{L}_{cts}$ & $\mathcal{L}_{contra}\mathcal{L}_{ass}$& $\mathcal{L}_{ass}\mathcal{L}_{cts}$ & 
        $\mathcal{L}_{contra}\mathcal{L}_{cts}$ &
        $\mathcal{L}_{contra}\mathcal{L}_{ass}\mathcal{L}_{cts}$
        \\
        \midrule
        PBT50RS & 86.1 & 86.5 & 84.3 & 86.6 & 86.9 & 86.4 & \bf87.1 \\
        \bottomrule
    \end{tabular}
    }
    \vspace{-1mm}
\end{table}

\begin{table}[!t]
    \centering
    \small
    \tabcolsep 4pt
    \caption{Ablation on clustering numbers.}
    \vspace{-1mm}
    \label{tab:ab_cluster_number}
    \scalebox{0.99}{
    \begin{tabular}{c|cccccc} 
    \toprule
    \#Cluster & 2 & 4 & 8 & 16 & 24 & 32 \\ 
    \midrule
    PBT50RS   & 85.46  & 86.32 & 86.36 & 86.71 & \bf87.12 & 86.99\\
    \bottomrule
    \end{tabular}
    }
    \vspace{-1mm}
\end{table}

\noindent \textbf{Number of clusters.}
We began by evaluating the impact of varying the number of groups in our method, adjusting $K$ from 2 to 32. As shown in Tab.~\ref{tab:ab_cluster_number}, classification performance on the ScanObjectNN PBT50RS subset improves as the number of clusters increases, peaking at 24 clusters with an accuracy of 87.12\%. Beyond this, performance stabilizes, indicating that 24 clusters provide an optimal balance between granularity and feature richness.

\noindent\textbf{Comparison of pretraining on ShapeNet and ScanNet.}
To evaluate the effect of different pretraining datasets, we pre-trained \ourmethod{} on both ShapeNet (SH) and ScanNet (SC). Tab.~\ref{tab:sh_sc_compare} reports detection performance in terms of AP${25}$ and AP${50}$, comparing our method with two strong baselines--MaskPoint and Point-BERT--both pre-trained on ScanNet.
Our method achieves the highest accuracy when pre-trained on ScanNet, reaching 64.4 AP$_{25}$ and 41.8 AP$_{50}$, which surpasses all competing methods. Even when pre-trained on ShapeNet, \ourmethod{} still outperforms both MaskPoint and Point-BERT, demonstrating strong generalization across object-centric and scene-centric datasets. These results highlight the robustness of our approach and confirm that our pretraining strategy effectively leverages diverse data sources for downstream tasks. More ablation studies and detailed analysis are provided in the \textit{Supp. Mat.}.
\begin{table}[!t]
\small
\renewcommand{\arraystretch}{0.5}
\caption{Comparison of pretraining datasets. Performance of \ourmethod{} pre-trained on ShapeNet (SH) and ScanNet (SC).}\label{tab:sh_sc_compare}
    \tabcolsep 3pt
    \vspace{-1mm}
    \resizebox{\columnwidth}{!}{%
    \begin{tabular}{cc|cc|cc|cc|cc} 
        \toprule
        \multicolumn{2}{c|}{Scratch} & \multicolumn{2}{c|}{\ourmethod(SH)} & \multicolumn{2}{c|}{\ourmethod(SC)}  & \multicolumn{2}{c|}{MaskPoint (SC)} & \multicolumn{2}{c}{Point-BERT (SC)}\\
        \midrule
        AP25$\uparrow$ & AP50$\uparrow$ & AP25$\uparrow$ & AP50$\uparrow$ & AP25$\uparrow$ & AP50$\uparrow$ & AP25$\uparrow$ & AP50$\uparrow$ & AP25$\uparrow$ & AP50$\uparrow$\\ 
        \midrule
        62.1 & 37.9 & \underline{63.9} & \underline{41.4} & \bf64.4 & \bf41.8 & 63.4 & 40.6 & 61.0 & 38.3\\ 
        \bottomrule
    \end{tabular}
    }
    \vspace{-1mm}
\end{table}

%% file: tabs/scanobjectnn.tex
\begin{table}[t!]
    \centering
    \small
    \setlength\tabcolsep{2.5pt}
    \setlength{\extrarowheight}{0.5pt}
    \caption{\textbf{Classification} results on ScanObjectNN. \texttt{DA} indicates the use of rotation data augmentation during fine-tuning. The overall accuracy (OA, \%) is reported.}
    \vspace{-1mm}
    \label{tab:scanobjectnn}
    \scalebox{0.91}{
    \begin{tabular}{lccccc}
    \toprule[0.95pt]
    \rotatebox{30}{Method} & \rotatebox{30}{\#Params(M)} & \rotatebox{30}{\texttt{DA}} & \rotatebox{30}{OBJBG} & \rotatebox{30}{OBJONLY} & \rotatebox{30}{PBT50RS}\\
    \midrule[0.6pt]
    \multicolumn{6}{c}{\textit{with standard ViTs and single-modal} ({\scshape Full})}\\
    \midrule[0.6pt]
    Scratch & 22.1 & $\times$ & 79.86 & 80.55 & 77.24\\
    OcCo~\cite{wang2021unsupervised} & 22.1 & $\times$ & 84.85 & 85.54 & 78.79\\
    Point-BERT~\cite{PointBERT} & 22.1 & $\times$ & 87.43 & 88.12 & 83.07\\
    MaskPoint~\cite{liu2022masked} & 22.1 & $\times$ & 89.30 & 88.10 & 84.30\\
    Point-MAE~\cite{pang2022masked} & 22.1 & $\times$ & 90.02 & 88.29 & 85.18 \\
    Point-CMAE~\cite{ren2024bringing} & 22.1 & $\times$ & 90.02 & {88.64} & {85.95}\\
    \rowcolor{linecolor}\ourmethod (ours) & 22.1 & $\times$ & \bf90.88 & \bf90.36 & \bf87.12\\
    \midrule
    Point-CMAE~\cite{ren2024bringing} & 22.1 & $\checkmark$ & {93.46} & {91.05} & 88.75\\
    PointClustering~\cite{long2023pointclustering} & 22.1 & $\checkmark$ & - & - & 87.30 \\
    PointGPT-S~\cite{chen2024pointgpt} & 29.2 & $\checkmark$ & 93.39 & 92.43 & 89.17 \\
    \rowcolor{linecolor}\ourmethod{~} (ours) & 22.1 & $\checkmark$ & \bf93.59 & \bf92.04 & \bf89.28 \\
    \midrule[0.6pt]
    \multicolumn{6}{c}{\textit{with hierarchical ViTs/multi-modal/post-process} ({\scshape Full})} \\
    \midrule[0.6pt]
    Point-M2AE~\cite{zhang2022point} & 15.3 & $\times$  & 91.22 & 88.81 & 86.43\\
    Joint-MAE~\cite{guo2023joint} & - & $\times$ & 90.94 & 88.86 & 86.07 \\
    ACT~\cite{dong2022autoencoders} & 22.1 & $\checkmark$  & 93.29 & 91.91 & 88.21\\
    \midrule[0.6pt]
    \multicolumn{6}{c}{\textit{with standard ViTs and single-modal} ({\scshape MLP-Linear})} \\
    \midrule[0.6pt]
    Point-MAE~\cite{pang2022masked} & 22.1 &  $\times$ & 82.58 & 83.52 & 73.08 \\
    Point-CMAE~\cite{ren2024bringing} & 22.1 & $\times$ & 83.48 & 83.45 & 73.15 \\
    \rowcolor{linecolor}\ourmethod{~} (Ours) & 22.1 & $\times$ & \bf84.18 & \bf84.34 & \bf76.79\\
    \midrule[0.6pt]
    \multicolumn{6}{c}{\textit{with standard ViTs and single-modal} ({\scshape MLP-3})} \\
    \midrule[0.6pt]
    Point-MAE~\cite{pang2022masked} & 22.1 & $\times$ & 84.29 & 85.24 & 77.34 \\
    Point-CMAE~\cite{ren2024bringing}  & 22.1 & $\times$ & 85.88 & 85.60 & 77.47 \\
    \rowcolor{linecolor}\ourmethod{~} (Ours) & 22.1 & $\times$ & \bf87.44 & \bf86.92 & \bf79.88 \\
    \bottomrule[0.95pt]
    \end{tabular}
    }
    \vspace{-1mm}
\end{table}

%% file: secs/5_conclusion.tex
\section{Conclusion}
\label{sec:conclusion}
This paper presented \ourmethod, a novel approach for unsupervised point cloud representation learning that leverages deep clustering within a ViT framework. \ourmethod{} captures both geometric structure and semantic content by integrating masked point modeling with clustering-based learning, effectively reconstructing dense semantic features. 
To enhance instance-level understanding, we introduced a global contrastive learning mechanism that contrasts different masked views of the same point cloud, encouraging ViTs to learn richer, more informative representations. 
Our findings underscore the potential of clustering-based methods in point cloud pretraining, providing a promising pathway for more robust and semantically meaningful 3D feature learning.

%% file: secs/X_suppl.tex
\clearpage
\setcounter{page}{1}
\setcounter{page}{1}

\maketitlesupplementary
\setcounter{section}{0}
\setcounter{figure}{0}    
\setcounter{table}{0}    

\renewcommand{\thetable}{\Alph{table}}
\renewcommand{\thefigure}{\Alph{figure}}

\section{Introduction}
In this supplementary material, we first provide a detailed explanation of the components of \ourmethod in \cref{supp:components}. Next, we delve into the technical aspects of masked clustering learning, including algorithmic details in \cref{supp:clustering}. Furthermore, we present additional results in \cref{supp:more_result}, such as classification and few-shot classification on the synthetic ModelNet40 dataset~\cite{wu20153d}, along with further ablation studies as referenced in the main paper. Finally, we include additional qualitative comparisons of object detection and part segmentation, showcasing the performance of models pre-trained with \ourmethod against state-of-the-art (SOTA) competitors (\cref{sec:sup_qua}), and conclude with a discussion on the broader impact in \cref{sec:impact}.

\section{More Details of \ourmethod{}}
In this section, we first introduce the details of Masked autoencoder (MAE) for point cloud with standard ViTs~\cite{vaswani2017attention,ren2023masked,dosovitskiy2020image}, and then we introduce more details about masked clustering learning. Lastly, we provide the implementation details of our proposed method.

\subsection{Masked autoencoder for point cloud ViTs}
\label{supp:components}
For the embedded tokens $\mathcal{T}=\{T_1, T_2,\cdots,T_N\}\in \mathbb{R}^{N \times C}$, to deploy the MAE strategy requires a mask $m$ applied to $\mathcal{T}$ and outputs the visible tokens $\mathcal{T}^{v} \in \mathbb{R}^{(1-r)N \times C}$ and the masked tokens $\mathcal{T}^{m} \in \mathbb{R}^{rN \times C}$, where $r$ denotes the mask ratio. Then, the MAE of the point cloud~\cite{PointBERT,pang2022masked} can be summarized as point tokens which are masked with random mask $m$ are fed into the encoder $f_{\varphi}(\cdot)$, and then the decoder $d_{\theta}(\cdot)$ predicts the features $\mathcal{F}_R$ of the masked center points $\mathcal{X}_m$ with distribution $\mathcal{D}$:
\begin{equation}
    \mathcal{F}_R=d_{\theta}\left(z\oplus f_{\phi}(\mathcal{X}_{m})\right), \quad z = f_{\varphi}(\mathcal{T}^{v}).
    \label{eq:eq1}
\end{equation}
Here $z \in \mathbb{R}^{(1-r)N \times C}$ denotes the latent feature of visible tokens $\mathcal{T}^{v}$, $f_{\phi}$ denotes the position embedding network. The symbol $\oplus$ denotes the concatenation of the token dimension. $\varphi,\phi$ and $\theta$ are the trainable parameters of the encoder and position embedding network, respectively.
In our case, two random masks, $m_a$ and $m_b$, are applied to the input data, producing the reconstructed features $\mathcal{F}^a_R$ and $\mathcal{F}^b_R$. For each mask, we set the mask ratio to 0.6, ensuring that 60\% of the input points are masked during the training process.

\subsection{More Details of Masked Clustering Learning}
\label{supp:clustering}
We employ a differentiable MinCut approach to learn token clustering, leveraging an affinity matrix to seamlessly integrate semantic and spatial relationships. This approach ensures that neighboring points tend to have strong similarities (whether based on spatial proximity or semantic features) and are consistently grouped into the same segment, enhancing the coherence and precision of the clustering.

By embedding both geometric and semantic information into the graph structure, this method effectively resolves ambiguities where points share similar features but differ in spatial context or object part associations. For instance, it can distinguish between different instances or parts of the same category, such as the two wings of an airplane.

Below, we provide additional details about Alg. 1, as presented in the main paper, focusing on three key aspects: \textit{Ensuring Connectivity}, \textit{Neighbor Weighting}, and \textit{Stabilizing Distance Weights}.

\begin{itemize}
    \item \textbf{Ensuring Connectivity}: To ensure clustering connectivity, we construct the affinity matrix $\mathcal{W}$ by considering only the $k$-nearest neighbors, where $k=4$ in our implementation. This approach encodes only local relationships, preventing the graph from becoming overly sparse or excessively dense, thereby maintaining a balance between computational efficiency and connectivity.

    \item \textbf{Neighbor Weighting}: To differentiate between neighboring points, we use the reconstructed feature ($\mathcal{F}_R \in \{\mathcal{F}^a_R, \mathcal{F}^b_R\}$) similarity to weight the connections. Specifically, the weights are computed as:
    \begin{equation*}
        \mathcal{C} \leftarrow 1 + \mathcal{F}_{R} \cdot \mathcal{F}_{R}^\top,
    \end{equation*}
    as shown in Step 4 of Algorithm 1. Note that $\mathcal{F}_R$ is normalized to maintain consistency and prevent numerical instability, ensuring that the values of the elements in $\mathcal{C}$ range from 0 to 2, \ie, $\mathcal{C} \in [0, 2]$.

    \item \textbf{Stabilizing Distance Weights}: To enhance stability, we normalize the distance-based weights by applying the following transformation:
    \[
    \mathcal{W} \leftarrow \exp \left( \overline{\text{Dis}} - \text{Dis} \right) \odot \mathcal{W},
    \]
    where the normalized distance term $\overline{\text{Dis}}_i$ is computed as:
    \[
    \overline{\text{Dis}}_i = \sum_j \text{Dis}_{ij}.
    \]
    This balances the distance weights, preventing any single connection from dominating the clustering process.
\end{itemize}

By combining these steps, our approach provides a robust and scalable solution for clustering tokens based on both semantic and spatial information, enabling more accurate segmentation in complex 3D objects.

\subsection{Implementation details}
We follow the experimental setup of Point-MAE~\cite{pang2022masked}, using a standard point cloud transformer~\cite{qi2017pointnet} as the encoder. 
In the autoencoder configuration, the encoder consists of 12 Transformer blocks, while the decoder contains 4 ViT encoder blocks. Each Transformer block has a hidden dimension of 384 and 6 attention heads, with an MLP ratio set to 4.
The encoder is pre-trained with our \ourmethod{} on the ShapeNet dataset~\cite{chang2015shapenet}, which includes 57,448 synthetic objects across 55 categories. Following~\cite{pang2022masked,ren2024bringing}, each point cloud is randomly downsampled to 1,024 points. We use the official ShapeNet training split for pre-training.
\ourmethod{} was implemented in PyTorch and trained on one Tesla V100-PCI-E-32G GPU. We trained the unsupervised representation learning model for 300 epochs using the AdamW optimizer~\cite{gugger2018adamw}, with a batch size of 128. 
The initial learning rate was set to 0.0005 and followed a cosine decay schedule with a decay weight of 0.05. During training, we set  $K = 24$  and  $\epsilon = 5 \times 10^{-4}$, as these values performed well in practice.

\section{More Results \& Discussion}
\label{supp:more_result}
We first present the classification and few-shot classification results on the synthetic ModelNet40 dataset~\cite{wu20153d}, as outlined in the main paper.

\input{tabs/modelnetcls}
\subsection{More results on ModelNet40~\cite{wu20153d}}
\paragraph{Object Classification.}
We evaluate 3D shape classification on the synthetic ModelNet40 dataset~\cite{wu20153d}, which includes 12,311 clean 3D CAD models spanning 40 object categories. Following the standard protocol~\cite{PointBERT,pang2022masked}, we divide the dataset into 9,843 training instances and 2,468 testing instances. Data augmentation during training involves random scaling and translation. 
The classification results, presented in Tab.~\ref{tab:modelnet}, use the voting strategy for fair comparison when reproducing Point-MAE~\cite{pang2022masked} and \ourmethod{} under the Full evaluation protocol. \ourmethod{} achieves competitive performance with an overall accuracy (OA) of $93.8\%$, matching Point-MAE while being slightly lower than PointClustering's $94.5\%$. Under the MLP-Linear and MLP-$3$ protocols, where the backbone remains frozen, \ourmethod{} surpasses both Point-MAE and Point-CMAE, with notable improvements in accuracy, showing strong generalization of its learned features.
Key findings include:
(i) Under the Full protocol, \ourmethod{} effectively captures informative representations and demonstrates comparable performance to SOTA methods on ModelNet40.
(ii) Under the MLP-Linear protocol, \ourmethod{} achieves the highest accuracy at $92.67\%$, outperforming Point-MAE and Point-CMAE by $1.45\%$ and $0.37\%$, respectively.
(iii) Under the MLP-$3$ protocol, \ourmethod{} delivers the best performance at $93.40\%$, further highlighting its robustness in transfer learning tasks.

\input{tabs/fewshot}
\noindent \textbf{Few-shot classification.} We evaluate \ourmethod~on the few-shot classification task using the ModelNet40 dataset~\cite{wu20153d}, following the established protocols from prior works~\cite{pang2022masked,PointBERT,liu2022masked,qi2023contrast}. The evaluation is conducted under a $num\_{cls}$-way, $num\_{sample}$-shot setting, where $num\_{cls}$ represents the number of randomly sampled classes, and $num\_{sample}$ denotes the number of objects sampled per class. For training, $num\_{cls} \times num\_{sample}$ objects are used, while testing involves 20 unseen objects from each of the $num\_{cls}$ classes.

We perform experiments with $num\_{cls} = \{5, 10\}$ and $num\_{sample} = \{10, 20\}$, reporting the mean accuracy and standard deviation over 10 independent runs to ensure statistical reliability. Importantly, no voting strategies are applied in this setting. Tab.~\ref{tab:few-shot} presents the results.
Our method, \ourmethod{}, achieves notable improvements over supervised baselines. For example, compared to models trained from scratch, it achieves improvements of +8.8\%, +4.7\%, +8.1\%, and +5.9\% across the four settings. Furthermore, \ourmethod{} demonstrates competitive or superior performance compared to state-of-the-art self-supervised approaches. Despite utilizing a standard ViT architecture pre-trained exclusively on point cloud data, \ourmethod{} matches or surpasses methods employing more complex architectures (\eg, Point-M2AE~\cite{zhang2022point}) or leveraging multimodal information (\eg, Joint-MAE~\cite{guo2023joint}, Point-GPT~\cite{chen2024pointgpt}, and ACT~\cite{dong2022autoencoders}).
For example, in the 5-way 20-shot setting, \ourmethod{} achieves an accuracy of 98.3\%, which is highly competitive and matches Point-M2AE\cite{zhang2022point}, despite using a simpler architecture. Similarly, in the 10-way 10-shot setting, it achieves an accuracy of 92.9\%, surpassing most state-of-the-art methods, including Joint-MAE~\cite{guo2023joint} and Point-GPT~\cite{chen2024pointgpt}, and closely matching Point-M2AE. These results highlight the robustness and efficiency of \ourmethod{} in handling challenging few-shot learning scenarios, achieving near state-of-the-art performance without relying on hierarchical or multimodal designs.

These results underline the capability of \ourmethod{} to balance simplicity and efficiency while achieving competitive performance in few-shot classification.

\subsection{Ablation Studies on ShapeNetPart~\cite{chang2015shapenet}}
\begin{table*}[t]
\centering
\setlength{\tabcolsep}{12pt}
\caption{Ablation study on ShapeNetPart~\cite{chang2015shapenet} for part segmentation. Mean intersection over union (mIoU) is reported for all classes (\textit{mIoU\textsubscript{C}}) and all instances (\textit{mIoU\textsubscript{I}}). We compare different clustering methods: MinCut, k-means, clustering without feature weighting (w/o feature), and a self-attention mechanism replacing GCN. MinCut demonstrates superior performance by effectively integrating semantic and geometric information.}
\label{tab:ab_kmeans}
\begin{tabular}{l cccccccc}
\toprule
Method & \multicolumn{2}{c}{MinCut} & \multicolumn{2}{c}{K-means} & \multicolumn{2}{c}{w/o feature} & \multicolumn{2}{c}{Self-attention} \\
\cmidrule(lr){2-3}\cmidrule(lr){4-5}\cmidrule(lr){6-7}\cmidrule(lr){8-9}
Metric & \textit{mIoU\textsubscript{C}} & \textit{mIoU\textsubscript{I}} & \textit{mIoU\textsubscript{C}} & \textit{mIoU\textsubscript{I}} & \textit{mIoU\textsubscript{C}} & \textit{mIoU\textsubscript{I}} & \textit{mIoU\textsubscript{C}} & \textit{mIoU\textsubscript{I}} \\
\midrule
Results & \bf85.2 & \bf86.4 & 84.1 & 85.5 & 84.2 & 85.8 & 83.6 & 85.2 \\
\bottomrule
\end{tabular}
\end{table*}
We conduct ablation studies on the ShapeNetPart dataset~\cite{chang2015shapenet} to evaluate the impact of different clustering methods on part segmentation performance. Tab.~\ref{tab:ab_kmeans} reports the mean intersection over union (mIoU) scores for all classes (\textit{mIoU\textsubscript{C}}) and all instances (\textit{mIoU\textsubscript{I}}). The following configurations are compared:
\begin{itemize}
    \item MinCut: Our proposed method that effectively integrates semantic and geometric information.
    \item K-means: A widely used clustering algorithm.
    \item w/o feature: Clustering without incorporating feature similarity weights.
    \item Self-attention: Replacing the GCN with a self-attention mechanism and MLP projector.
\end{itemize}

As shown in Tab.~\ref{tab:ab_kmeans}, the MinCut approach consistently achieves the highest performance for both \textit{mIoU\textsubscript{C}} and \textit{mIoU\textsubscript{I}}, demonstrating its superiority in capturing semantic and geometric relationships. Incorporating feature similarity weights yields significant improvements compared to setups without feature weighting, underscoring the importance of feature integration for clustering accuracy.
Conversely, replacing GCN with a self-attention mechanism and an MLP projector, followed by a softmax operation (as described in Eq.~(2) of the main paper), results in a slight performance drop. This indicates that the GCN is more effective at capturing local dependencies, which are crucial for learning segment-level semantic features in 3D point cloud segmentation.

These results underscore the robustness of the MinCut approach in learning a soft partitioning function that effectively groups point clouds into semantically coherent clusters. By seamlessly integrating both semantic and geometric information, MinCut proves essential for achieving strong performance in 3D shape segmentation. Furthermore, the study highlights the pivotal role of the GCN in preserving local connectivity and semantic relationships, which are crucial for accurate and meaningful part segmentation.

\subsection{MAE and clustering integration challenges.}
Simply combining these two paradigms in a multitask learning setup yields lower performance than the MAE-based baseline (reproduced 93.6). The main reason is that MAE-based approaches learn \emph{pose-aware} features, whereas clustering-based methods tend to learn \emph{pose-invariant} features. 
Hence, directly combining them introduces a conflict. 
We verified this by extracting point-level features from non-rotated and randomly rotated versions of each point cloud (with rotation angles in \([45^\circ, 90^\circ]\)) using encoders pre-trained by PointMAE and by \ourmethod{} on ModelNet40. 
After applying max-pooling to obtain global features, we computed the similarity between the non-rotated and rotated versions.
The table above shows that features pretrained via clustering achieve a higher average similarity than those learned through MAE, suggesting that the clustering-based approach generates more pose-invariant features.
\begin{table}[!t]
    \centering
    \tabcolsep 2pt
    \caption{Performance and pose sensitivity comparison between MAE and clustering-based pretraining. Clustering achieves higher feature similarity under rotation, indicating stronger pose invariance, but underperforms in classification when combined with MAE.}
    \vspace{-2mm}
    \scalebox{1.0}{
    \begin{tabular}{cc|cc}
        \toprule
        \multicolumn{2}{c|}{Classification} & \multicolumn{2}{c}{Similarity} \\
        \midrule
        PointMAE & MAE+Clustering & PointMAE & Clustering\\
        \midrule
        93.6 & 92.9 & 0.74 & 0.93\\
        \bottomrule
    \end{tabular}%
    }
\end{table}

\subsection{Efficiency Comparison}
Compared to other MAE-based methods, \ourmethod{} introduces only a modest increase in both parameters and GFLOPs, primarily due to the lightweight GCN used in the clustering branch, while maintaining high efficiency during pretraining, as shown in Tab.~\ref{supp:efficiency}. Specifically, \ourmethod{} adds just 2.7M parameters and 1.1 GFLOPs over Point-MAE, and remains significantly more efficient than recent counterparts such as Point-FEMAE and Recon, which incur noticeably higher computational costs. This demonstrates that our design strikes a favorable balance between effectiveness and efficiency, enabling scalable pretraining without relying on heavy architectures.

\begin{table}[!t]
\renewcommand{\arraystretch}{0.5} 
\tabcolsep 2pt
\caption{The Efficiency Comparison.}
\vspace{-2mm}
\label{supp:efficiency}
\scalebox{0.6}{
    \begin{tabular}{cc|cc|cc|cc} 
        \toprule
        \multicolumn{2}{c|}{Point-MAE (Baseline)} & \multicolumn{2}{c|}{Point-FEMAE} & \multicolumn{2}{c|}{Recon} & \multicolumn{2}{c}{\ourmethod} \\ 
        \#Params (M) & GFLOPS & \#Params (M) & GFLOPS & \#Params (M) & GFLOPS & \#Params (M) & GFLOPS \\ 
        \midrule
        29.0 & 2.1 & 41.5 & 4.7 & 140.9 & 19.2 & 31.7 & 3.2 \\ 
        \bottomrule
    \end{tabular}
}
\end{table}

\section{More qualitative results}
\label{sec:sup_qua}
In this section, we provide additional qualitative results to showcase the effectiveness of \ourmethod{} in addressing key 3D understanding tasks, including object detection and part segmentation. The visualizations compare our method against baseline approaches and ground truth annotations, demonstrating its robustness, accuracy, and capability to handle complex 3D environments and intricate object structures. These results further validate the superiority of \ourmethod{} in real-world scenarios.
\begin{figure}[!t]
    \centering
    \includegraphics[width=1\linewidth]{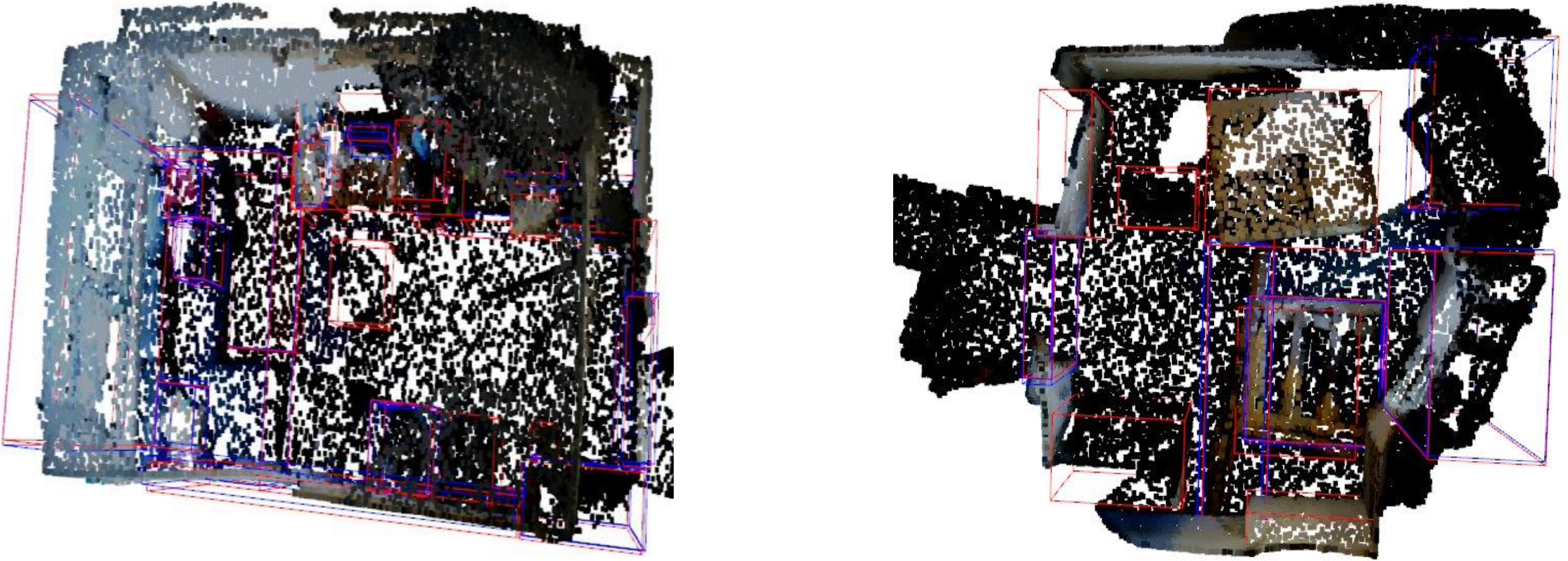}
    \caption{Visualization of detection results on 2 more examples from the ScanNet dataset. Red boxes indicate the ground truth annotations, while blue boxes represent the model’s predictions. These visualizations showcase the model’s capability to handle challenging 3D environments with varying spatial structures and object distributions, demonstrating the robustness and effectiveness of the detection approach.}
    \label{fig:sup_detection}
\end{figure}

\noindent\textbf{Detection results.} Fig.~\ref{fig:sup_detection} presents detection results on six examples from the ScanNet dataset~\cite{dai2017scannet}. Red boxes indicate the ground truth annotations, while blue boxes represent the model’s predictions. These examples highlight the model's ability to accurately detect and localize objects in complex 3D environments, demonstrating strong alignment between predictions and ground truth.

\begin{figure*}[!t]
    \centering
    \begin{overpic}[width=1.0\linewidth]{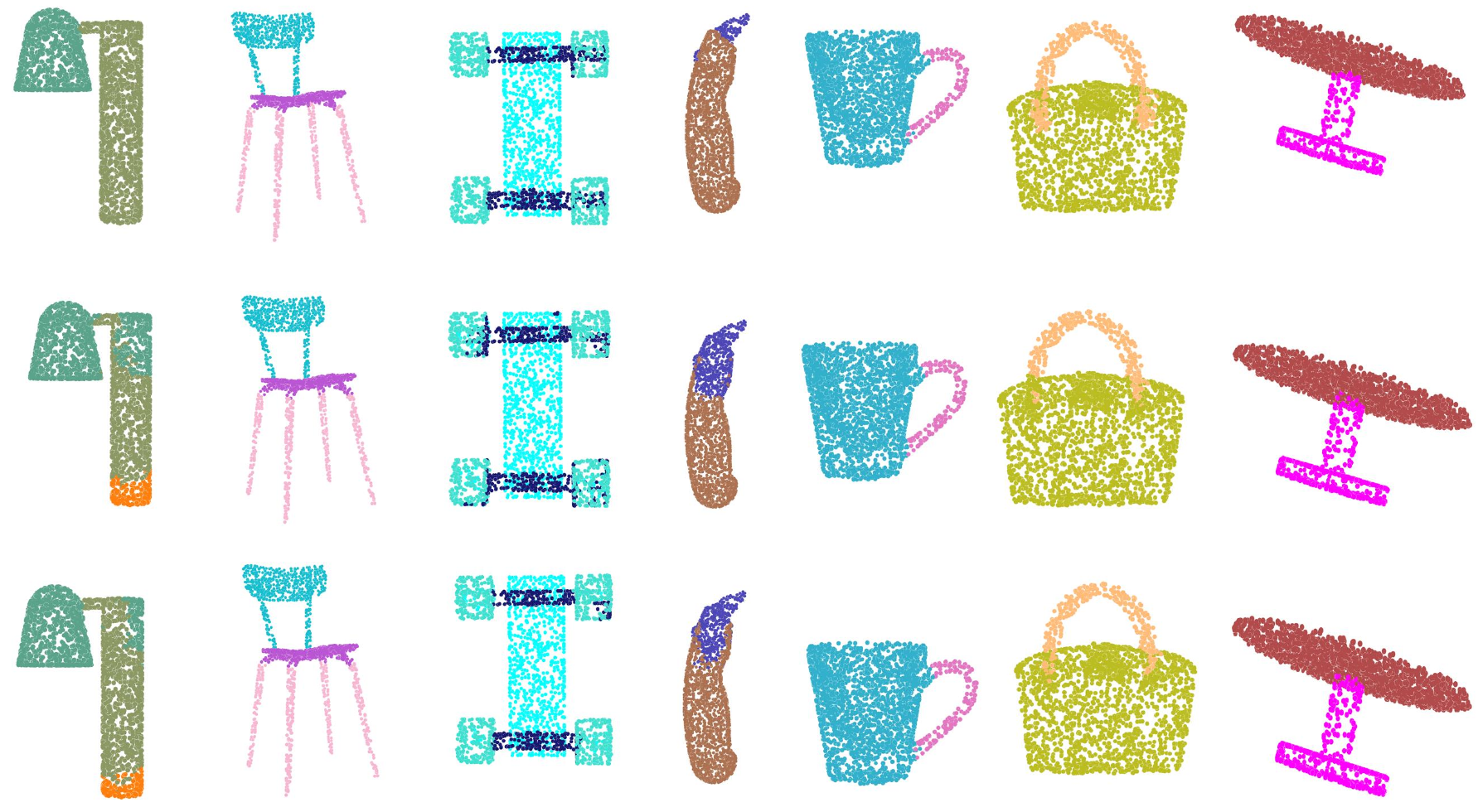}
        \put(-0.5,45){\color{black}\footnotesize \rotatebox{90}{\textbf{GT}}}
        \put(-0.5,25){\color{black}\footnotesize \rotatebox{90}{Point-MAE}}
        \put(-0.5,5){\color{black}\footnotesize \rotatebox{90}{\textbf{\ourmethod{} (Ours)}}}
    \end{overpic}
    \caption{Part segmentation results on the ShapeNetPart dataset~\cite{chang2015shapenet}. The visualization compares the predictions from \ourmethod{} and PointMAE~\cite{chen2023pimae} against the ground truth annotations (\textbf{GT}). The examples illustrate the superior ability of \ourmethod{} to segment 3D objects into their constituent parts with high accuracy, capturing intricate details of object structures.}
    \label{fig:partseg_results}
\end{figure*}

In this section, we present additional qualitative results to further demonstrate the effectiveness of \ourmethod{} in various 3D understanding tasks, including part segmentation and object detection. The visualizations highlight the robustness and accuracy of \ourmethod{} compared to baseline methods and ground truth annotations. These results showcase the capability of our method to handle complex 3D scenes and intricate object structures, providing strong evidence of its performance in real-world scenarios.

\noindent\textbf{Part segmentation.} We present the part segmentation results of \ourmethod{} on the ShapeNetPart dataset~\cite{chang2015shapenet}. The visualization highlights a comparison between our method (\textbf{\ourmethod}), the baseline method Point-MAE~\cite{pang2022masked}, and the ground truth annotations (\textbf{GT}). These results demonstrate the effectiveness of \ourmethod{} in accurately segmenting 3D objects into their constituent parts, outperforming existing approaches in capturing fine-grained object details. 

\begin{figure*}[!t]
    \centering
    \includegraphics[width=1\linewidth]{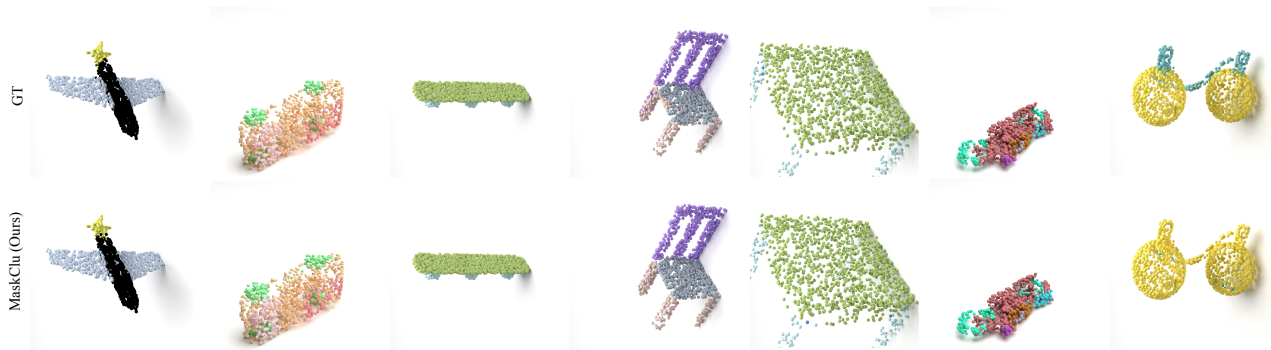}
    \caption{Visualization of part segmentation results between our method and the ground truth.}
    \label{fig:partseg_more}
\end{figure*}

Besides the comparison to Point-MAE, we also provide more comparison samples between our method and ground truth in Fig.~\ref{fig:partseg_more}. 
The visual comparison shows that our method can produce super close part segmentation results compared to the GT, which further validates the effectiveness of our \ourmethod{}. 
Notably, the last example in Fig.\ref{fig:partseg_more} also illustrates a failure case, where the connection between the two bicycle wheels is not correctly segmented.

\section{Limitation \& Future Work}
\label{supp:limittation}
\ourmethod{} achieves competitive performance on ModelNet40 and clearly outperforms prior methods under linear probing and lightweight MLP evaluation protocols, demonstrating strong generalization of the learned features. Although its performance under the Full protocol is marginally lower than PointClustering, we believe this is partially due to the clean and synthetic nature of ModelNet40, where clustering-based features tend to excel by capturing pose-invariant features.
In contrast, our method shows significant advantages on real-world datasets such as ScanObjectNN (Please refer to Tab.~3 of our main manuscript), where it consistently surpasses existing approaches across various settings. This highlights the robustness and practicality of our learned representations in noisy and incomplete scenarios, validating the effectiveness of combining masked modeling and clustering.

In future work, we aim to further explore adaptive strategies to balance pose-aware and pose-invariant cues, potentially through disentangled representation learning or dynamic task modulation. 
We also plan to extend \ourmethod{} to more challenging 3D understanding tasks, such as part segmentation or open-set recognition in real-world scenes.

\section{Broader Impact}
\label{sec:impact}
The proposed framework, \ourmethod{}, has the potential to significantly advance the field of 3D point cloud understanding, with far-reaching implications across various domains. By enabling Vision Transformers (ViTs) to learn richer and more semantically meaningful representations, \ourmethod{} could contribute to breakthroughs in applications such as autonomous driving, robotics, medical imaging, and urban planning. For instance, improved point cloud segmentation and classification could enhance object detection in autonomous vehicles, facilitate precise surgical navigation in medical robotics, and support efficient city modeling for smart urban development.
Moreover, the unsupervised nature of \ourmethod{} reduces reliance on large-scale annotated datasets, which are often costly and labor-intensive to produce. This accessibility fosters broader adoption of point cloud processing techniques, particularly in resource-constrained environments or less-explored domains where labeled data may be scarce. By making the code and pre-trained models publicly available, we aim to democratize access to cutting-edge 3D representation learning tools, encouraging collaboration and innovation in both academia and industry.
However, like any machine learning approach, \ourmethod{} is not without potential risks. Enhanced 3D understanding could be misused in surveillance, military, or other ethically sensitive applications, raising concerns about privacy and misuse of technology. To mitigate such risks, we advocate for responsible research and application of our method, guided by ethical standards.
Overall, \ourmethod{} represents a step forward in 3D vision research, broadening the scope of practical applications while promoting sustainable, ethical, and accessible technological progress.

%% file: tabs/modelnetcls.tex
\begin{table}[!t]
    \centering
    \small
    \caption{Classification results on the ModelNet40 dataset, with overall accuracy (OA, \%) reported. \texttt{[ST]} denotes the standard Transformer. $^{*}$ indicates reproduced results.}\label{tab:modelnet}
    \setlength\tabcolsep{6pt}
    \scalebox{1.0}{
    \begin{tabular}{lccc}
    \toprule[0.81pt]
    Method & \texttt{[ST]} & \#Point & OA (\%)\\
    \midrule[0.6pt]
    \multicolumn{4}{c}{\textit{Supervised learning only}}\\
    \midrule[0.6pt]
    PointNet~\cite{qi2017pointnet} & - & 1k P & 89.2\\
    PointNet++~\cite{qi2017pointnet++} & - & 1k P & 90.7\\
    PointCNN~\cite{li2018pointcnn} & - & 1k P & 92.5\\
    DGCNN~\cite{wang2019dynamic} & - & 1k P & 92.9\\
    DensePoint~\cite{liu2019densepoint} & - & 1k P & 93.2\\
    PointASNL~\cite{wu2019pointconv} & - & 1k P & 92.9\\
    DRNet~\cite{qiu2021dense} & - & 1k P & 93.1\\
    \midrule[0.6pt]
    Point Trans.~\cite{engel2021point} & $\times$ & 1k P & 92.8\\
    PCT~\cite{guo2021pct} & $\times$ & 1k P & 93.2\\
    PointTransformer~\cite{zhao2021point} & $\times$ & 1k P & 93.7\\
    NPCT~\cite{guo2021pct} & $\checkmark$ & 1k P & 91.0\\
    \midrule[0.6pt]
    \multicolumn{4}{c}{\textit{with self-Supervised learning} ({\scshape Full})}\\
    \midrule[0.6pt]
    Scratch & $\checkmark$ & 1k P & 91.4\\
    OcCo~\cite{wang2021unsupervised} & $\checkmark$ & 1k P & 92.1\\
    Point-BERT~\cite{PointBERT} & $\checkmark$ & 1k P & 93.2\\
    Point-MAE~\cite{pang2022masked} & $\checkmark$ & 1k P & 93.8\\
    Point-MAE$^{*}$~\cite{pang2022masked} & $\checkmark$ & 1k P &93.5\\
    Point-CMAE~\cite{ren2024bringing} & $\checkmark$ & 1k P & 93.6\\
    PointClustering~\cite{long2023pointclustering} & $\checkmark$ & 1k P & \bf{94.5}\\
    PointClustering$^{*}$\cite{long2023pointclustering} & $\checkmark$ & 1k P & 93.4\\
    \rowcolor{linecolor}\ourmethod(ours) & $\checkmark$ & 1k P & 93.8\\
    \midrule[0.6pt]
    \multicolumn{4}{c}{\textit{with self-supervised Learning} ({\scshape Mlp-Linear})}\\
    \midrule[0.6pt]
    Point-MAE~\cite{pang2022masked} & $\checkmark$ & 1k P & 91.22$\pm$0.26\\
    Point-CMAE~\cite{ren2024bringing} & $\checkmark$ & 1k P & 92.30$\pm$0.32\\
    \rowcolor{linecolor}\ourmethod(ours) & $\checkmark$ & 1k P & \textbf{92.67}$\pm$\textbf{0.29}\\
    \midrule[0.6pt]
    \multicolumn{4}{c}{\textit{with self-supervised Learning} ({\scshape Mlp-$3$})}\\
    \midrule[0.6pt]
    Point-MAE~\cite{pang2022masked} & $\checkmark$ & 1k P & 92.33$\pm$0.09\\
    Point-CMAE~\cite{ren2024bringing} & $\checkmark$ & 1k P & 92.60$\pm$0.19 \\
    \rowcolor{linecolor}\ourmethod(ours) & $\checkmark$ & 1k P & \textbf{93.40}$\pm$\textbf{0.12}\\
    \bottomrule[0.99pt]
    \end{tabular}
    }
\end{table}

%% file: tabs/fewshot.tex
\begin{table*}[t!]
    \centering
    \small
    \caption{\textbf{Few-shot classification} results on the ModelNet40 dataset. The table reports overall accuracy (\%) for 5-way and 10-way tasks under 10-shot and 20-shot settings. Results are grouped by methods using supervised representation learning, self-supervised ViTs, and hierarchical/multi-modal/self-supervised ViTs. \ourmethod~achieves competitive performance, consistently outperforming many baselines and demonstrating state-of-the-art accuracy in several settings. Best results are shown in bold, and second-best results are underlined.}
    \label{tab:few-shot}
    \setlength\tabcolsep{20pt}
    \begin{tabular}{lcccc}
    \toprule[0.95pt]
    \multirow{2}{*}[-0.5ex]{Method}& \multicolumn{2}{c}{{5-way}} & \multicolumn{2}{c}{{10-way}}\\
    \cmidrule(lr){2-3}\cmidrule(lr){4-5}
    & 10-shot & 20-shot & 10-shot & 20-shot\\
    \midrule[0.6pt]
    \multicolumn{5}{c}{\textit{Supervised Representation Learning}}\\
    \midrule[0.6pt]
    PointNet~\cite{li2018point} &52.0 $\pm$ 3.8 &  57.8 $\pm$ 4.9&  46.6 $\pm$  4.3& 35.2 $\pm$ 4.8 \\
    PointNet-OcCo~\cite{wang2021unsupervised} &89.7 $\pm$ 1.9 &  92.4 $\pm$ 1.6 &  83.9 $\pm$ 1.8 & 89.7 $\pm$ 1.5\\
    PointNet-CrossPoint~\cite{afham2022crosspoint} &90.9 $\pm$ 4.8 &  93.5 $\pm$ 4.4 &  84.6 $\pm$ 4.7 & 90.2 $\pm$ 2.2\\
    DGCNN~\cite{wang2019dynamic} &31.6 $\pm$ 2.8 &  40.8 $\pm$ 4.6&  19.9 $\pm$  2.1& 16.9 $\pm$ 1.5\\
    DGCNN-CrossPoint~\cite{afham2022crosspoint} &92.5 $\pm$ 3.0 & 94.9 $\pm$ 2.1 &83.6 $\pm$ 5.3 &87.9 $\pm$ 4.2\\
    \midrule[0.6pt]
    \multicolumn{5}{c}{\textit{ViTs with self-supervision} ({\scshape Full})}\\
    \midrule[0.6pt]
    Scratch & 87.8${\pm}$5.2& 93.3${\pm}$4.3 & 84.6${\pm}$5.5 & 89.4${\pm}$6.3\\
    OcCo \cite{wang2021unsupervised} & 94.0${\pm}$3.6& 95.9${\pm}$2.3 & 89.4${\pm}$5.1 & 92.4${\pm}$4.6\\
    Point-BERT \cite{PointBERT} & 94.6${\pm}$3.1 & 96.3${\pm}$2.7 & 91.0${\pm}$5.4 & 92.7${\pm}$5.1\\
    MaskPoint \cite{liu2022masked} & 95.0${\pm}$3.7 & 97.2${\pm}$1.7 & 91.4${\pm}${4.0} & 93.4${\pm}$3.5\\
    Point-MAE \cite{pang2022masked} & 96.3${\pm}$2.5& {97.8}${\pm}$1.8 & {92.6}${\pm}$4.1 & {95.0}${\pm}${3.0}\\
    Point-CMAE \cite{ren2024bringing} & \bf{96.7}${\pm}${2.2} & \underline{98.0${\pm}${0.9}} & \underline{92.7${\pm}$4.4} & \bf{95.3}${\pm}$3.3 \\
    \rowcolor{linecolor}\ourmethod{} (Ours) & \underline{96.5${\pm}$1.6} & \bf98.3${\pm}$1.1 & \bf92.9${\pm}$3.3 & \underline{95.2${\pm}$2.3} \\
    \midrule[0.6pt]
    \multicolumn{5}{c}{\textit{ViTs with hierarchical/multi-modal/self-supervision} ({\scshape Full})}\\
    \midrule[0.6pt]
    Point-M2AE \cite{zhang2022point} & 96.8${\pm}$1.8 & 98.3${\pm}$1.4 & 92.3${\pm}$4.5 & 95.0${\pm}$3.0 \\
    Joint-MAE \cite{guo2023joint} & 96.7${\pm}$2.2 & 97.9${\pm}$1.8 & 92.6${\pm}$3.7 & 95.1${\pm}$2.6 \\
    PointGPT-B \cite{chen2024pointgpt} & 97.5${\pm}$2.0 & 98.8${\pm}$1.0 & 93.5${\pm}$4.0 & 95.8${\pm}$3.0 \\
    ACT \cite{dong2022autoencoders} & 96.8${\pm}$2.3 & 98.0${\pm}$1.4 & 93.3${\pm}$4.0 & 95.6${\pm}$2.8 \\
    \bottomrule[0.95pt]
    \end{tabular}
\end{table*}